\documentclass{article}

\usepackage{PRIMEarxiv}

\usepackage[utf8]{inputenc} 
\usepackage[T1]{fontenc}    
\usepackage{hyperref}       
\usepackage{url}            
\usepackage{booktabs}       
\usepackage{amsfonts}       
\usepackage{nicefrac}       
\usepackage{microtype}      
\usepackage{lipsum}
\usepackage{fancyhdr}       
\usepackage{graphicx}       
\graphicspath{{media/}}     

\usepackage{amsmath}
\usepackage{amssymb}
\usepackage{babel}
\usepackage{multirow,booktabs}
\usepackage{tabularx}
\usepackage{graphicx}
\usepackage{epstopdf} 
\usepackage{multicol}
\usepackage[linesnumbered]{algorithm2e}
\usepackage{enumerate}
\usepackage{rotating}
\usepackage{fixltx2e}
\usepackage[utf8]{inputenc}
\usepackage{xcolor}
\usepackage{adjustbox}
\newcommand\hl[1]{%
	\bgroup
	\hskip0pt\color{red!80!black}%
	#1%
	\egroup
}
\usepackage{wrapfig}

\usepackage{color, colortbl}
\definecolor{LightRed}{rgb}{1,0.5,0.35}

\usepackage{tikz}
\usetikzlibrary{fit,calc}

\usepackage{lineno,hyperref}
\modulolinenumbers[5]
\usepackage{graphicx} 
\usepackage{epstopdf}    

\title{Clustering based on Mixtures of Sparse Gaussian Processes}

\author{
  Zahra Moslehi\\
  Isfahan University of Technology\\
  Isfahan, Iran \\
  \texttt{z.moslehi@ec.iut.ac.ir, zahra.moslehi@dkfz-heidelberg.de} \\
   \And
  Abdolreza Mirzaei, Mehran Safayani \\
  Isfahan University of Technology\\
  Isfahan, Iran \\
  \texttt{\{mirzaei, safayani\}@cc.iut.ac.ir} \\
}

\begin{document}
\maketitle

\begin{abstract}
Creating low dimensional representations of a high dimensional data set is an important component in many machine learning applications. How to cluster data using their low dimensional embedded space is still a challenging problem in machine learning. In this article, we focus on proposing a joint formulation for both clustering and dimensionality reduction. When a probabilistic model is desired, one possible solution is to use the mixture models in which both cluster indicator and low dimensional space are learned. Our algorithm is based on a mixture of sparse Gaussian processes, which is called Sparse Gaussian Process Mixture Clustering (SGP-MIC). The main advantages to our approach over existing methods are that the probabilistic nature of this model provides more advantages over existing deterministic methods, it is straightforward to construct non-linear generalizations of the model, and applying a sparse model and an efficient variational EM approximation help to speed up the algorithm.
\end{abstract}


\section{Introduction}
Machine learning is basically categorized into supervised and unsupervised learning. One approach in unsupervised learning is finding the low-dimensional embedded space to better represent the structure of all input data. Calculating similarity measures in high-dimensional data suffers from high computational complexity. Moreover, distances in low-dimensional space are more meaningful than their raw representation. Thus, dimensionality reduction and feature transformation methods have been the focus of many researchers these years. There exist some deterministic and probabilistic approaches to learn such low-dimensional space. For example, Principal Component Analysis (PCA) \cite{jolliffe1986principal} and ISOMAP \cite{tenenbaum2000global} are two linear and non-linear traditional deterministic methods and Probabilistic PCA (PPCA) \cite{tipping1999probabilistic} and Gaussian Process Latent Variable Model (GP-LVM) \cite{lawrence2004gaussian,lawrence2005probabilistic} are two linear and non-linear probabilistic methods.

In this paper, a dimensionality reduction algorithm with real-world application in clustering is proposed. Each cluster must contain the same class label and the separability and compactness on all clusters have to be maximized. One way to achieve better clustering result is to perform a dimensionality reduction method such as PCA or PPCA as the preprocessing methods and then cluster data points in this low-dimensional embedded space. Although, these dimensionality reduction methods can capture the intrinsic structure of input data, but applying them before clustering may not help to find the best clusters. A better way is to consider the requirements of clustering during the process of dimensionality reduction and vice versa. In this paper, we focus on finding the low-dimensional space and the optimal set of clusters simultaneously and in a joint formulation.

Combining Linear Discriminant Analysis (LDA) to define the low dimensional embedded space and k-means to cluster data is proposed by Ding \textit{et al.} \cite{ding2002adaptive,ding2007adaptive}. Discriminative Cluster Analysis (DCA) as another joint model is proposed by Torre and Kanade, in which the eigenvalue decomposition is applied to define the linear transformation matrix and a gradient based method is applied to define the cluster indicator matrix \cite{de2006discriminative}. Following these works, Adaptive Metric Learning (AML) algorithm is proposed by Ye \textit{et al.}, in which instead of the gradient based method, Kernel k-means is applied for clustering \cite{ye2007adaptive}. The simpler formulation to remove the alternating behavior of previous methods leads to the Discriminative K-means (DisKmeans) algorithm, proposed by Ye \textit{et al.} \cite{ye2008discriminative}. The nonlinear version of AML named NAML is also investigated by Chen \textit{et al.} \cite{chen2007nonlinear}. Optimized Kernel K-means Clustering (OKKC) with lower computational complexity than NAML is then proposed by Yu \textit{et al.} \cite{yu2012optimized}. Unsupervised Neighborhood Component Analysis (UNCA) is another algorithm, which is based on the regularized NCA and k-means clustering \cite{qin2015unsupervised}. Similarity-based Discriminative Clustering (SDC) applies regularized techniques to both increase the inter-cluster separability while avoids the pitfalls of highly discriminative methods like collapsing or over-fitting the representation to the noisy data \cite{passalis2019discriminative}. Discriminative Embedded Clustering (DEC), unlike methods which consist of a combination of supervised metric learning and k-means algorithms, combines unsupervised method PCA with k-means \cite{hou2014discriminative}. Discriminative Fuzzy C-Means (Dis-FCM) integrates linear dimensionality reduction and fuzzy clustering FCM in a joint formulation \cite{moslehi2018discriminative}. Its probabilistic version is Bayesian Discriminative Fuzzy C-Means (BDFC) \cite{heidari2018bayesian}. Feature-Reduction FCM (FRFCM) is another fuzzy clustering method which learns the feature weights and eliminates the features with small weight to improve the complexity and performance of FCM method \cite{yang2017feature}. Recently, \textit{deep clustering} methods using deep neural networks have been proposed in learning both clustering and deep representation \cite{min2018survey}. The loss function in these methods is composed of two main terms, network loss and clustering loss. The network loss can be the reconstruction loss of an autoencoder (AE)  \cite{ji2017deep}, \cite{yang2017towards}, the variational loss of a variational autoencoder (VAE), \cite{jiang2017variational}, or the adversarial loss of a generative adversarial network (GAN) \cite{chen2016infogan} and the clustering loss can be the loss function of k-means clustering or any other clustering method. The network and clustering losses in AE-based methods are independent, thus there is no theoretical analysis on explaining why they improve the clustering performance. VAE-based deep clustering methods, as the generative variant of AE, have the good theoretical guarantee since they minimize the variational lower bound on the marginal likelihood of data points, but they have the problem of high-computational complexity. Hard convergence and mode collapse are also the problems of GAN-based methods \cite{min2018survey}. There exists another category in deep clustering in which the network loss is removed. They are referred as clustering deep neural network (CDNN-based) methods where they have the risk of learning a corrupted feature representation \cite{hsu2017cnn}, \cite{xie2016unsupervised}. The clustering loss should be defined carefully in these methods and network initialization is very important here. Moreover, a major problem of all deep clustering methods is that they are mostly designed for handling image data sets and they are not working on a wide range of different data types. Having the sufficient data is another need for deep algorithms. They have lots of hyper parameters where their best values need to be defined by lots of tuning. The way to design kernels is tricky in them. Also, their network structure has also lots of effect on the results. Among these methods, AE-based, GAN-based and CDNN-based methods are in deterministic area and VAE-based methods are in probabilistic perspective. Moreover, mixture models have been developed to learn both clustering and low-dimensional space in a probabilistic manner \cite{chazan2019deep}, \cite{uugur2020variational}. A mixture model can be interpreted as a dimensionality reduction and clustering mixing model in which each component is considered to be a specific cluster. The advantages of probabilistic models over deterministic approaches are that in probabilistic modeling the dependency relationships among all different variables can be represented by a diagrammatic data structure, called probabilistic graphical models; and the information about different latent variables are applied to define the prior distributions \cite{bishop2006machine}. Capturing uncertainty and decreasing time complexity are the other benefits of probabilistic models.

Concentrating on probabilistic modeling, we propose a new mixture model to cluster data in the low dimensional embedded space. Our model is based on mixture dual probabilistic PCA (PPCA) proposed in Gaussian Process Latent Variable Model (GP-LVM) \cite{lawrence2004gaussian, lawrence2005probabilistic}. The manner in which the mixture model is constructed allows for non-linearization of the each component. Our mixture of DPPCA model can be interpreted as a mixture of GPs. However, there are lots of mixture of GPs models \cite{li2021mixture, gadd2020enriched, luo2017variational, mcdowell2018clustering} but there are very few mixture of GPs in the context of clustering, in which the input space is latent. In our proposed method, we use the sparse approximation suggested in \cite{csato2002sparse,seeger2003fast} to speed up the model. Sparse GP is not a new idea but mixing it with the cluster indicator parameter to be able to learn both clustering and low dimensional embedded data is a new idea which is followed in this paper. This model is solved through an efficient variational inference approach leading to a practical algorithm, named \textit{Sparse Gaussian Process MIxture Clustering} (\textit{SGP-MIC}).
Briefly, the features specific to this proposed method are summarized as follows:
\begin{itemize}
	\item SGP-MIC introduces a new framework based on mixture of sparse GP-LVM to cluster data points in the low dimensional embedded space. Its probabilistic aspect has lots of benefits than other deterministic ones.
	\item It is straightforward to ‘non-linearize’ the model. Deep mixture models and VAE-based deep clustering are relevant methods. Recently, quite a few publications have shown that GP and DL are equivalent \cite{garriga2018infiniteconv}. They have demonstrated that wide network is equivalent to GP as well.
	\item An efficient variational EM approximation is proposed in SGP-MIC to alleviate the poor speed of convergence problem that exists in standard variational EM approximation.
\end{itemize}

The remainder of this paper is organized as follows: All mathematical details about SGP-MIC are described in Section \ref{sec:SGP-MIC}; We will present experimental results obtained from the model on a range of different data sets in Section \ref{sec:Experiments}, and finally  the article is concluded in Section \ref{sec:Conclusion}.

\section{Sparse Gaussian Process Mixture Clustering (\textit{SGP-MIC}) \label{sec:SGP-MIC}}
In this section, we first review the Gaussian Process (GP) and then introduce our proposed model.

\subsection{A Brief Review on Gaussian Processes \label{sec:GP}}
Gaussian Processes (GPs) are a kind of probabilistic models where define a distribution over functions. In GPs, a function is viewed as an infinite dimensional vector, where a prior distribution is defined on a set of $N$ instances of them. This prior distribution is Gaussian parametrized by a mean and a covariance. The mean is considered to be zero and the covariance is a function of input space on which the process operates. The covariance computes the similarity between all pairs from the input space, which is the inverse of a distance measure. By sampling from the GP prior distribution, when a pair of input data points are close together, their function values are high correlated. Thus, it generates a smooth function over the input space. When the input space is considered as a latent variable, it is known as the GP Latent Variable Model (GP-LVM) \cite{lawrence2004gaussian, lawrence2005probabilistic}.
\subsection{Probabilistic Model \label{sec:ProbabilisticModel}}
First we explain briefly the main idea of our probabilistic model. We assume that the observed data in high dimensional space $\mathbf{Y}$ is the noisy version of true values $\mathbf{F}$. We define a nonlinear function between $\mathbf{F}$ as the data in high dimensional space and $\mathbf{X}$ in low dimensional space. This function is defined by a Gaussian Process (GP) while the input space $\mathbf{X}$ in this GP is latent. The idea of latent input space is coming from the GP-LVM algorithm in which they suppose the features are independent, but the samples are highly correlated. This GP can capture the correlation between data points in high dimensional space by learning $\mathbf{X}$ in such a way that if two data points in high dimensional space are high correlated, they must be close together in low dimensional space. In this way, we will be able to extract the manifold structure of data points. What is different from the main GP-LVM algorithm and makes the formulation more complicated is that here we are considering the sparse GP by defining some inducing points and at the same time defining an extra variable $\mathbf{S}$, which is the cluster indicator. We can interpret this model as a sparse GP-LVM by splitting the latent space into subspace to learn better manifold structure of data points. Here we use the variable $\mathbf{S}$ to define the membership values of each data to each cluster. That is why we claim that in our method we are learning both clusters and low dimensional embedded space in a joint formulation. In the following, we explain the probability distribution and  dependency between different variables.

Assume that $\mathbf{Y}\in\Re^{N\times P}$ is the given data set, where $N$ is the number of data points and $P$ is the number of attributes. Here, we will introduce a mixture model containing $M$ separate components. Let $\mathbf{S}\in\left\{0,1\right\}^{N\times M}$ be the indicator matrix to determine which data point is represented by which component. This matrix indicates that if $n$-th data point belongs to the $m$-th component, $s_{n,m}$ is set to one while all other values are set to zero. The prior distribution of this indicator matrix is the multinomial distribution,
\begin{equation}
	p\left(\mathbf{S}|\boldsymbol{\pi}\right)=\prod_{n=1}^{N}\prod_{m=1}^{M}\pi_{m}^{s_{n,m}},
\end{equation}
where $\boldsymbol{\pi}$ is a $M$ dimensional vector in which the prior probability of the $m$-th component of the mixture is defined by $\pi_{m}$ and
\begin{equation*}
	\sum_{m=1}^{M}\pi_{m}=1.
\end{equation*}

In real world application, it is typical to assume that observed data points $\mathbf{Y}$ are noisy measurements of true values $\mathbf{F}$. This induces the factorized likelihood takes the form:
\begin{equation}
	\begin{aligned}[b]
		& p\left(\mathbf{Y}|\mathbf{S},\left\{ \mathbf{F}^{\left(m\right)}\right\} _{m=1}^{M},\beta\right)\\
		& =\prod_{n=1}^{N}\prod_{m=1}^{M}\prod_{i=1}^{P} \mathcal{N} \left(y_{n,i}|f_{i}^{\left(m\right)}\left(\mathbf{x}_{n}\right),\beta^{-1}\right)^{s_{n,m}},
	\end{aligned}
\end{equation}
where $f_{i}^{\left(m\right)}\left(\mathbf{x}_{n}\right)$ is a function of a $Q$-dimensional latent space associated with the $m$-th component of the mixture. The function values for each component $m$ are denoted by $\mathbf{F}^{\left(m\right)}\in\Re^{N\times P}$, where $f_{n,i}^{\left(m\right)}=f_{i}^{\left(m\right)}\left(\mathbf{x}_{n}\right)$. The points in lower dimensional latent space are represented through the matrix $\mathbf{X}\in\Re^{N\times Q}$. The relationship between the latent space and the data space is given by the function $f_{i}^{\left(m\right)}\left(\mathbf{x}_{n}\right)$. We choose a sparse Gaussian process prior for $\left\{ \mathbf{F}^{\left(m\right)}\right\}$. In sparse GPs, the idea is to expand the probability space with $N'$ different auxiliary pairs of input-output variables collected in matrices $\mathbf{X}_{u}\in\Re^{N'\times Q}$ and $\mathbf{U}^{\left(m\right)}\in\Re^{N'\times P}$. Here, the inducing output variables $\mathbf{U}^{\left(m\right)}$ are assumed to be in a same GP prior with variables $\mathbf{F}^{\left(m\right)}$. The factorized prior over these variables therefore takes the form, 
\begin{equation}
	\begin{aligned}[b]
		& p\left(\left\{ \mathbf{F}^{\left(m\right)}\right\} _{m=1}^{M},\left\{\mathbf{U}^{\left(m\right)}\right\} _{m=1}^{M}|\mathbf{X},\mathbf{X}_{u},\boldsymbol{\theta}\right)\\
		& 
		=\prod_{m=1}^{M}\prod_{i=1}^{P}\mathcal{N}\left(
		\begin{bmatrix}\mathbf{f}_{:,i}^{\left(m\right)}\\
			\mathbf{u}_{:,i}^{\left(m\right)}
		\end{bmatrix}|\mathbf{0},
		\begin{bmatrix}\mathbf{K}_{ff}^{\left(m\right)} & \mathbf{K}_{fu}^{\left(m\right)}\\
			\mathbf{K}_{uf}^{\left(m\right)} & \mathbf{K}_{uu}^{\left(m\right)}
		\end{bmatrix}
		\right)
		,
	\end{aligned}
\end{equation}
where $\mathbf{K}_{ff}$ is built by computing the covariance function on all latent variables $\mathbf{X} $, $\mathbf{K}_{uu}$ is constructed by evaluating the covariance function on all auxiliary samples $\mathbf{X}_{u}$, $\mathbf{K}_{fu}$ is cross covariance between latent variables and auxiliary samples, and $\mathbf{K}_{uf}=\mathbf{K}_{fu}^{T}$. The dependence on variables $\mathbf{X}$, $\mathbf{X}_{u}$, and the parameters, $\boldsymbol{\theta}$, is through these kernel matrices.
GP definition allows us to write the marginal distribution and conditional distributions as follows \cite{Rasmussen:2005:GPM:1162254}:
\begin{equation}
	\begin{aligned}[b]
		p\left(\left\{\mathbf{U}^{\left(m\right)}\right\}|\mathbf{X}_{u}\right)=
		\prod_{m=1}^{M}\prod_{i=1}^{P}\mathcal{N}\left(\mathbf{u}_{:,i}^{\left(m\right)}|\mathbf{0},\mathbf{K}_{uu}^{\left(m\right)}
		\right),
	\end{aligned}
	\label{eq:MarginU}
\end{equation}

\begin{equation}
	\begin{aligned}[b]
		& p\left(\left\{\mathbf{F}^{\left(m\right)}\right\}|\left\{\mathbf{U}^{\left(m\right)}\right\} ,\mathbf{X},\mathbf{X}_{u}, \boldsymbol{\theta}\right)\\
		& =\prod_{m=1}^{M}\prod_{i=1}^{P}\mathcal{N}\left(\mathbf{f}_{:,i}^{\left(m\right)}|\mathbf{a}_{:,i}^{\left(m\right)},\tilde{\mathbf{K}}^{\left(m\right)}
		\right),\\
		\mathbf{a}_{:,i}^{\left(m\right)}&=\mathbf{K}_{fu}^{\left(m\right)}{\mathbf{K}_{uu}^{\left(m\right)}}^{-1}\mathbf{u}_{:,i}^{\left(m\right)},\\
		\tilde{\mathbf{K}}^{\left(m\right)}&=\mathbf{K}_{ff}^{\left(m\right)}-\mathbf{K}_{fu}^{\left(m\right)}{\mathbf{K}_{uu}^{\left(m\right)}}^{-1}\mathbf{K}_{uf}^{\left(m\right)}.
		\label{eq:CondF|U}
	\end{aligned}
\end{equation}
By using equations (\ref{eq:MarginU}) and (\ref{eq:CondF|U}), the marginal distribution
\begin{align}
	& p\left(\left\{\mathbf{F}^{\left(m\right)}\right\}|\mathbf{X},\mathbf{X}_{u}, \boldsymbol{\theta}\right)\nonumber\\
	& =\prod_{m=1}^{M}\prod_{i=1}^{P}\int p\left(\mathbf{f}_{:,i}^{\left(m\right)}|\mathbf{u}_{:,i}^{\left(m\right)},\mathbf{X},\mathbf{X}_{u}, \boldsymbol{\theta}\right)p\left(\mathbf{u}_{:,i}^{\left(m\right)}|\mathbf{X}_{u}\right)d\mathbf{u}_{:,i}^{\left(m\right)}\nonumber\\
	& =\prod_{m=1}^{M}\prod_{i=1}^{P} \mathcal{N}\left(\mathbf{f}_{:,i}^{\left(m\right)}|\mathbf{0},\underbrace{\mathbf{K}_{ff}^{\left(m\right)}-\mathbf{K}_{fu}^{\left(m\right)}{\mathbf{K}_{uu}^{\left(m\right)}}^{-1}\mathbf{K}_{uf}^{\left(m\right)}}_{\mathrm{\tilde{\mathbf{K}}^{\left(m\right)}}}\right.\nonumber\\
	& \left.\; \; \; \; \; \; \; \; \; \; \; \; \; \; \; \; \; \; \; \; \; \; \; \; \; \; \; \; \; \; \; \; \; \; \; \; \; \; \; \; \; \; \;+\mathbf{K}_{fu}^{\left(m\right)}{\mathbf{K}_{uu}^{\left(m\right)}}^{-1}\mathbf{K}_{uf}^{\left(m\right)}\right).
	\label{eq:MarginF}
\end{align}
Using an approximate posterior, we have  
\begin{equation}
	\begin{aligned}[b]
		\tilde{p}\left(\mathbf{f}_{:,i}^{\left(m\right)}|\mathbf{u}_{:,i}^{\left(m\right)},\mathbf{X},\mathbf{X}_{u},\boldsymbol{\theta}\right)=\mathcal{N}\left(\mathbf{f}_{:,i}^{\left(m\right)}|\mathbf{a}_{:,i}^{\left(m\right)},\tilde{\mathbf{Q}}^{\left(m\right)}
		\right),
	\end{aligned}
\end{equation}
where $\tilde{\mathbf{Q}}^{\left(m\right)}\neq\tilde{\mathbf{K}}^{\left(m\right)}$. Here, we applied deterministic training conditional (DTC) approximation with $\tilde{\mathbf{Q}}^{\left(m\right)}=0$ \cite{csato2002sparse,seeger2003fast}. Thus, the marginal distribution will be: 
\begin{align}
	& \tilde{p}\left(\left\{\mathbf{F}^{\left(m\right)}\right\} |\mathbf{X},\mathbf{X}_{u},\boldsymbol{\theta}\right)\nonumber\\
	& =\prod_{m=1}^{M}\prod_{i=1}^{P} \mathcal{N}\left(\mathbf{f}_{:,i}^{\left(m\right)}|\mathbf{0},\mathbf{K}_{fu}^{\left(m\right)}{\mathbf{K}_{uu}^{\left(m\right)}}^{-1}\mathbf{K}_{uf}^{\left(m\right)}\right).
\end{align}
This sparse GP model is associated with a computational cost of $O\left(NN'^{2}\right)$, $N'\ll N$ \cite{damianou2015deep}.
Finally, we apply a prior distribution across the latent space, 
\begin{equation}
	p\left(\mathbf{X}|\mathbf{S}\right)=\prod_{n=1}^{N}\prod_{m=1}^{M}\mathcal{N}\left(\mathbf{x}_{n}|\bar{\mathbf{x}}_{m},\mathbf{C}_{m}\right)^{s_{n,m}}.
\end{equation}

\begin{figure}[t]
	\centering	
	\includegraphics[scale=1]{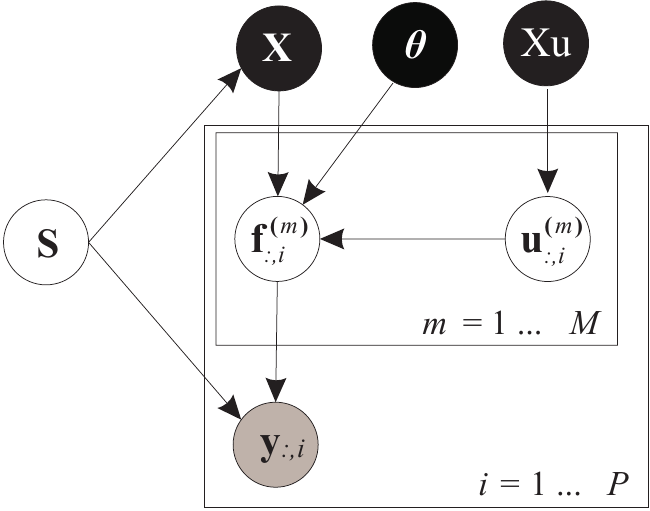} 
	\caption{Probabilistic graphical model for SGP-MIC.}\footnotesize
	\label{figure:PGM}
\end{figure}

Figure \ref{figure:PGM} graphically illustrates this proposed probability model. In this graphical model, shaded and white color nodes represent observed and latent variables respectively; black color circles represent parameters where they need to be optimized by derivation of likelihood function. This model is effectively a mixture of sparse Gaussian processes. This sparse mixture model is an extension of a preliminary work suggested by Urtasun and Lawrence \cite{MGPLVM}. Exact inference in a mixture of GPs is computationally intractable \cite{tresp2001mixtures}, and thus variational approximations are proposed in the next section.

\subsection{Standard Variational Approximation\label{sec:variationalApprox}}
In standard variational approximation, given observed data $\mathbf{D}$ and the latent variables $\mathbf{H}$, the data log likelihood is written as $\ln (\mathbf{D})=\mathcal{L_{SV}}(q)+\mbox{KL}(q||p)$, where $\mathcal{L_{SV}}(q)=\int q(\mathbf{H})\log{\left \{\frac{p(\mathbf{H},\mathbf{D})} {q(\mathbf{H})}\right \}} d\mathbf{H}$ and Kullback-Leibler divergence $\mbox{KL}(q||p)$ is a nonnegative value. $\mathcal{L_{SV}}(q)$ is a lower bound of marginal log likelihood which should be maximized \cite{jordan1999introduction}.

In this model, the variational posterior distribution is assumed to take the factorized form:
\begin{equation}
	q\left(\left\{ \mathbf{F}^{\left(m\right)}\right\},\mathbf{S}\right)=q\left(\left\{ \mathbf{F}^{\left(m\right)}\right\}\right)q\left(\mathbf{S}\right).
\end{equation}
By this factorization some additional factorizations fall out of the variational posteriors. The posterior approximation for $\mathbf{S}$ factorizes w.r.t data points and the components of the mixture,
\begin{equation}
	q\left(\mathbf{S}\right)=\prod_{n=1}^{N}\prod_{m=1}^{M}q\left(s_{n,m}\right)\label{eq:q_S}
\end{equation}
and $q\left(\left\{\mathbf{F}^{\left(m\right)}\right\}\right)$ is factorized as follows:
\begin{equation}
	q\left(\left\{ \mathbf{F}^{\left(m\right)}\right\} \right)=\prod_{m=1}^{M}\prod_{i=1}^{P}q\left(\mathbf{f}_{:,i}^{\left(m\right)}\right).\label{eq:q_F}
\end{equation}
This implies that lower bound of marginal log likelihood is found as:
\begin{align}
	& \mathcal{L_{SV}}\left(\mathbf{X},\mathbf{X}_{u},\boldsymbol{\theta},\beta,\boldsymbol{\pi},\left\{\bar{\mathbf{x}}_{m},\mathbf{C}_{m}\right\}\right)\nonumber\\
	& = \left\langle\log p\left(\mathbf{Y}|\left\{\mathbf{F}^{\left(m\right)}\right\} ,\mathbf{S},\beta\right)\right.\nonumber\\
	& \left. \quad \quad \quad \; \tilde{p}\left(\left\{ \mathbf{F}^{\left(m\right)}\right\} |\mathbf{X},\mathbf{X}_{u},\boldsymbol{\theta}\right)\right\rangle_{q\left(\left\{ \mathbf{F}^{\left(m\right)}\right\} \right)q(\mathbf{S})}\nonumber\\
	& + \left\langle\log p\left(\mathbf{X}|\mathbf{S}\right) p\left(\mathbf{S}|\boldsymbol{\pi}\right)\right\rangle _{q(\mathbf{S})} \nonumber\\  
	& -\sum_{n=1}^{N}\sum_{m=1}^{M}\left\langle \log q\left(s_{n,m}\right)\right\rangle _{q\left(s_{n,m}\right)}\nonumber\\  
	& -\sum_{m=1}^{M}\sum_{i=1}^{P}\left\langle \log q\left(\mathbf{f}_{:,i}^{\left(m\right)}\right)\right\rangle _{q\left(\mathbf{f}_{:,i}^{\left(m\right)}\right)}. \label{eq:traditionalBound1}
\end{align}
Substituting our model probability distributions, we obtain 
\begin{align}
	& \mathcal{L_{SV}}\left(\mathbf{X},\mathbf{X}_{u},\boldsymbol{\theta},\beta,\boldsymbol{\pi},\left\{\bar{\mathbf{x}}_{m},\mathbf{C}_{m}\right\}\right)\nonumber \\ 
	& =\sum_{n=1}^{N}\sum_{m=1}^{M}\sum_{i=1}^{P} \left\langle s_{n,m} \log \mathcal{N}\left(y_{n,i}|f_{n,i}^{\left(m\right)},\beta^{-1}\right)\right\rangle _{q\left(\mathbf{f}_{:,i}^{\left(m\right)}\right)q\left(s_{n,m}\right)}\nonumber \\
	& +\sum_{m=1}^{M}\sum_{i=1}^{P}\left<\log N\left(\mathbf{f}_{:,i}^{\left(m\right)}|\mathbf{0},\mathbf{K}_{fu}^{\left(m\right)}{\mathbf{K}_{uu}^{\left(m\right)}}^{-1}\mathbf{K}_{uf}^{\left(m\right)}\right)\right>_{q\left(\mathbf{f}_{:,i}^{\left(m\right)}\right)}\nonumber \\
	& +\sum_{n=1}^{N}\sum_{m=1}^{M}\left\langle s_{n,m}\right\rangle_{q\left(s_{n,m}\right)} \log \mathcal{N}\left(\mathbf{x}_{n}|\bar{\mathbf{x}}_{m},\mathbf{C}_{m}\right)\nonumber \\
	& +\sum_{n=1}^{N}\sum_{m=1}^{M}\left\langle s_{n,m}\right\rangle_{q\left(s_{n,m}\right)} \log\pi_{m}\nonumber \\
	& -\sum_{n=1}^{N}\sum_{m=1}^{M}\left\langle \log q\left(s_{n,m}\right)\right\rangle_{q\left(s_{n,m}\right)}\nonumber\\  
	& -\sum_{m=1}^{M}\sum_{i=1}^{P}\left\langle \log q\left(\mathbf{f}_{:,i}^{\left(m\right)}\right)\right\rangle_{q\left(\mathbf{f}_{:,i}^{\left(m\right)}\right)}.
	\label{eq:traditionalBound}
\end{align}

An EM algorithm can be applied to maximize the above lower bound. The EM-style algorithm alternates between  updating the form of each variational approximation and the maximizing (\ref{eq:traditionalBound}) with respect to the model parameters $\left(\mathbf{X},\mathbf{X}_{u},\boldsymbol{\theta},\beta,\boldsymbol{\pi},\left\{\bar{\mathbf{x}}_{m},\mathbf{C}_{m}\right\}\right)$.
By equating the functional derivatives of Equation (\ref{eq:traditionalBound}) with respect to $q(.)$ to zero, the variational posteriors $q\left(s_{n,m}\right)$ and $q\left(\mathbf{f}_{:,i}^{\left(m\right)}\right)$ are derived as follows (\ref{appendix:qs}, \ref{appendix:qf}):

\begin{equation}
	\nonumber
	q\left(s_{n,m}\right)=\left(\frac{p_{n,m}}{\sum_{m=1}^{M}p_{n,m}}\right)^{s_{n,m}},
\end{equation}

\begin{equation}
	\begin{aligned}[b]
		p_{n,m}=&\pi_{m}\mathcal{N}\left(\mathbf{x}_{n}|\bar{\mathbf{x}}_{m},\mathbf{C}_{m}\right)\\
		& \exp\left(\left\langle \log\prod_{i=1}^{P}\mathcal{N}\left(y_{n,i}|f_{n,i}^{\left(m\right)},\beta^{-1}\right)\right\rangle_{q\left(\mathbf{f}_{:,i}^{\left(m\right)}\right)}\right).
		\label{eq:q_Snm}
	\end{aligned}
\end{equation}

\begin{equation}
	\begin{aligned}[b]
		q\left(\mathbf{f}_{:,i}^{\left(m\right)}\right) \propto \; &\mathcal{N}\left(\mathbf{f}_{:,i}^{\left(m\right)}|\mathbf{0},\mathbf{K}_{fu}^{\left(m\right)}{\mathbf{K}_{uu}^{\left(m\right)}}^{-1}\mathbf{K}_{uf}^{\left(m\right)}\right)\\
		& \mathcal{N}\left(\mathbf{y}_{:,i}|\mathbf{f}_{:,i}^{\left(m\right)},\left(\mathbf{B}^{\left(m\right)}\right)^{-1}\right),
	\end{aligned}
\end{equation}
where $\mathbf{B}^{\left(m\right)}\in\Re^{N\times N}$ is a diagonal matrix with elements $b_{n,n}^{\left(m\right)}=\beta\left\langle s_{n,m}\right\rangle $.
Another way to express this is
\begin{equation}
	\begin{aligned}[b]
		& q\left(\mathbf{f}_{:,i}^{\left(m\right)}\right)\propto \mathcal{N}\left(\mathbf{f}_{:,i}^{\left(m\right)}|\bar{\mathbf{f}}_{:,i}^{\left(m\right)},\mathbf{ \Sigma}\right), \bar{\mathbf{f}}_{:,i}^{\left(m\right)}=\mathbf{\Sigma}\mathbf{B}^{\left(m\right)}\mathbf{y}_{:,i}\\
		& \boldsymbol{\mathbf{\Sigma}}=\left(\left(\mathbf{K}_{fu}^{\left(m\right)}{\mathbf{K}_{uu}^{\left(m\right)}}^{-1}\mathbf{K}_{uf}^{\left(m\right)}\right)^{-1}+\mathbf{B}^{\left(m\right)}\right)^{-1}.
		\label{eq:q_fm}
	\end{aligned}
\end{equation}

A key problem of standard variational approximation is the slow speed of convergence. We therefore turn to an improved variational bound, known as KL-corrected bound proposed by King and Lawrence \cite{king2006fast}. This inference technique was proposed for solving a different probabilistic model, where its efficiency motivates us to apply it for solving our probabilistic model. This bound improves the speed of variational learning, without losing the guarantee of convergence. 
\subsection{KL-corrected Inference\label{sec:KLcorrected}}
The KL-corrected bound is a lower bound on the true likelihood, but an upper bound on the standard variational bound. To introduce this bound in our model, we first consider the marginal log likelihood,
\begin{align}
	& L\left(\mathbf{X},\mathbf{X}_{u},\boldsymbol{\theta},\beta,\boldsymbol{\pi},\left\{\bar{\mathbf{x}}_{m},\mathbf{C}_{m}\right\}\right)\nonumber \\
	& =\log\int \mbox{d}\left\{ \mathbf{F}^{\left(m\right)}\right\} p\left(\mathbf{Y}|\left\{\mathbf{F}^{\left(m\right)}\right\},\beta\right)p\left(\mathbf{X}\right)\nonumber \\
	& \; \; \; \; \; \; \; \; \; \; \; \; \; \; \; \; \; \; \; \; \; \; \; \; \; \; \; \; \; \; \tilde{p}\left(\left\{\mathbf{F}^{\left(m\right)}\right\} |\mathbf{X},\mathbf{X}_{u},\boldsymbol{\theta}\right). \label{eq:desiredMarginal}
\end{align}
Rather than lower bounding this likelihood, we only focus the term $p\left(\mathbf{Y}|\left\{\mathbf{F}^{\left(m\right)}\right\},\beta\right)p\left(\mathbf{X}\right)$ in the
integral where can be lower bounded variationally,
\begin{equation}
	\begin{aligned}[b]
		& p\left(\mathbf{Y}|\left\{\mathbf{F}^{\left(m\right)}\right\} ,\beta\right)p\left(\mathbf{X}\right)\\ 
		& \geq  \prod_{n=1}^{N}\prod_{m=1}^{M}\prod_{i=1}^{P}\exp\left( \right.\\
		& \left.\frac{1}{2}\left\langle s_{n,m}\right\rangle \log\frac{\beta}{2\pi} -\frac{\beta}{2}\left\langle s_{n,m}\right\rangle \left(y_{n,i}-f_{n,i}^{\left(m\right)}\right)^{2} \right)\\
		& \\
		& \times\prod_{n=1}^{N}\prod_{m=1}^{M} \exp\left( \left\langle s_{n,m}\right\rangle\log \mathcal{N}\left(\mathbf{x}_{n}|\bar{\mathbf{x}}_{m},\mathbf{C}_{m}\right) \right. \\
		& \left. \; \; \; \; \;  \; \; \; \; \; \; \; \; \; \; \; \; \; \; \; \; \; \; \; + \left\langle s_{n,m}\right\rangle \log\pi_{m} -\left\langle \log q\left(s_{n,m}\right)\right\rangle \right).
	\end{aligned}
\end{equation}
Substituting this bound into (\ref{eq:desiredMarginal}), we have
\begin{equation}
	\begin{aligned}[b]
		& L\left(\boldsymbol{\mathbf{X},\mathbf{X}_{u},\theta},\beta,\boldsymbol{\pi},\left\{\bar{\mathbf{x}}_{m},\mathbf{C}_{m}\right\}\right) \\ 
		& \geq\sum_{m=1}^{M}\sum_{i=1}^{P}\log\int\prod_{n=1}^{N}\mathcal{N}\left(y_{n,i}|f_{n,i}^{\left(m\right)},\left(\beta\left\langle s_{n,m}\right\rangle \right)^{-1}\right)\\
		& \; \; \; \; \; \; \; \; \; \; \; \; \; \; \; \; \; \; \; \; \; \; \; \; \; \; \; \mathcal{N}\left(\mathbf{f}_{:,i}^{\left(m\right)}|\mathbf{0},\mathbf{K}_{fu}^{\left(m\right)}{\mathbf{K}_{uu}^{\left(m\right)}}^{-1}\mathbf{K}_{uf}^{\left(m\right)}\right)\mbox{d}\mathbf{f}_{:,i}^{\left(m\right)}\\
		& +\sum_{n=1}^{N}\sum_{m=1}^{M} c_{n,m}\\
		& \doteq\mathcal{L}_{\mbox{KL}}\left(\mathbf{X},\mathbf{X}_{u},\boldsymbol{\theta},\beta,\boldsymbol{\pi},\left\{\bar{\mathbf{x}}_{m},\mathbf{C}_{m}\right\}\right),\label{eq:KLBound1}
	\end{aligned}
\end{equation}
where 
\begin{equation}
	\begin{aligned}[b]
		& c_{n,m}\\
		& =\left\langle s_{n,m}\right\rangle\log \mathcal{N}\left(\mathbf{x}_{n}|\bar{\mathbf{x}}_{m},\mathbf{C}_{m}\right)\\
		& +{\left\langle s_{n,m}\right\rangle }\log\pi_{m}-\left\langle \log q\left(s_{n,m}\right)\right\rangle \\
		& + P\times\log\sqrt{\frac{\left(\frac{\beta}{2\pi}\right)^{\left\langle s_{n,m}\right\rangle }}{\left\langle s_{n,m}\right\rangle \frac{\beta}{2\pi}}}.
	\end{aligned}
\end{equation}

The integral in this bound can now be computed analytically:

\begin{equation}
	\begin{aligned}[b]
		& \mathcal{L}_{\mbox{KL}}\left(\mathbf{X},\mathbf{X}_{u},\boldsymbol{\theta},\beta,\boldsymbol{\pi},\left\{\bar{\mathbf{x}}_{m},\mathbf{C}_{m}\right\}\right) \\
		& =\sum_{m=1}^{M}\sum_{i=1}^{P}\log \mathcal{N}\left(\mathbf{y}_{:,i}|\mathbf{0},\left(\mathbf{K}_{fu}^{\left(m\right)}{\mathbf{K}_{uu}^{\left(m\right)}}^{-1}\mathbf{K}_{uf}^{\left(m\right)}+{\mathbf{B}^{\left(m\right)}}^{-1}\right)\right)\\
		& +\sum_{n=1}^{N}\sum_{m=1}^{M} c_{n,m}, \label{eq:KLBound}
	\end{aligned}
\end{equation}
that equals the marginal likelihood when $q\left(s_{n,m}\right)=p\left(s_{n,m}|\mathbf{Y},\mathbf{X},\mathbf{X}_{u},\boldsymbol{\theta},\beta,\boldsymbol{\pi},\left\{\bar{\mathbf{x}}_{m},\mathbf{C}_{m}\right\}\right)$. The new bound (\ref{eq:KLBound}) is a lower bound on the true likelihood, but an upper bound on the standard variational bound (\ref{eq:traditionalBound}) (\ref{appendix:CorrectionTerm}). As such it is a more attractive candidate for optimization than the standard variational bound. Having true posterior distribution of the indicator matrix, the KL-corrected bound can be optimized with respect to the all parameters. Since computation of true posterior is analytically intractable, we apply the standard variational approximation given by (\ref{eq:q_Snm}) for updating $q\left(\mathbf{S}\right)$, which is combined with updates of $q\left(\left\{ \mathbf{F}^{\left(m\right)}\right\}\right)$. Our algorithm therefore proceeds as follows: In the E-step, $q\left(\mathbf{S}\right)$ is computed. The M-step then consists of maximization of the KL-corrected bound (\ref{eq:KLBound}) with respect to the parameters of $\mathbf{X},\mathbf{X}_{u},\boldsymbol{\theta},\beta,\boldsymbol{\pi},\left\{\bar{\mathbf{x}}_{m},\mathbf{C}_{m}\right\}$ using gradient based methods. We alternate between E-step and M-step to converge to the final solution. Algorithm 1 concludes all the steps, in which the implementation is done. Each gradient step of this algorithm requires an inverse of the kernel matrix, which has $O\left(NN'^{2}\right)$, $N'\ll N$ time complexity \cite{damianou2015deep}.

\IncMargin{1.5em}
\begin{algorithm}[t]	
	\caption{Implementation flow of the SGP-MIC.}  
	\SetKwInOut{Input}{input}
	\SetKwInOut{Output}{output}
	\Indm\Indmm
	\Input{Raw data points $\mathbf{Y}\in\Re^{N\times P}$, number of clusters $M$, reduced dimension $Q$, number of iterations $N_{iter}$.}
	\Output{$\mathbf{X}\in\Re^{N\times Q}$, $q\left(\mathbf{S}\right)$.}
	\Indp\Indpp
	\BlankLine
	Initialize $\mathbf{X},\mathbf{X}_{u},\boldsymbol{\theta},\beta,\boldsymbol{\pi},\left\{\bar{\mathbf{x}}_{m},\mathbf{C}_{m}\right\}$ according to the Section \ref{sec:Experimental Setting}\;
	\For{$N_{iter}$ iterations}{ 
		\textbf{E-step}: Update the values of $q\left(\left\{ \mathbf{F}^{\left(m\right)}\right\} \right)$ using  (\ref{eq:q_fm}) and then update the values of $q\left(\mathbf{S}\right)$ using (\ref{eq:q_Snm})\;
		\textbf{M-step}: Optimize (\ref{eq:KLBound}) with respect to the parameters of $\mathbf{X},\mathbf{X}_{u},\boldsymbol{\theta},\beta,\boldsymbol{\pi},\left\{\bar{\mathbf{x}}_{m},\mathbf{C}_{m}\right\}$ using scaled conjugate gradients.\; }
	\textbf{Return} $\mathbf{X}\in\Re^{N\times Q}$, $q\left(\mathbf{S}\right)$.
	\BlankLine
	\Indm\Indmm
	\Indp\Indpp
\end{algorithm}
\DecMargin{1.5em}

\section{Experiments\label{sec:Experiments}}

In this section, the effectiveness of SGP-MIC is evaluated through different experiments. In the following, first, the experimental setting and then the results are presented.

\begin{table}[b]
	\centering
	\caption{Summary of data sets applied in this article.}\footnotesize
		\begin{tabular}{ | c | c | c | c |}
			\hline
			\text{Data Set} & \text{Number of} & \text{Number of} & \text{Number of} \\ 
			& \text{Instances} & \text{Attributes} & \text{Classes} \\ \hline
			Iris & 150 & 4 & 3 \\ \hline
			Wine & 178 & 13 & 3 \\ \hline			
			Sonar & 208 & 60 & 2  \\ \hline
			WDBC & 569 & 30 & 2  \\ \hline
			Statlog-Australian & 690 & 14 & 2 \\ \hline
			Segment & 2310 & 19 & 7 \\ \hline
			Breast-cancer (BC) & 683 & 9 & 2 \\ \hline
			CMC & 1473 & 9 & 3 \\ \hline
			Yale & 165 & 1024 & 15 \\ \hline
		\end{tabular}
	\label{table:data sets}
\end{table}

\subsection{Experimental Setting\label{sec:Experimental Setting}}

Different experiments are run on 8 benchmark data sets of the UCI repository \cite{bache2013uci} and an image large scale data set Yale. The selected data sets have different number of data points, attributes, and classes where their details are given in Table \ref{table:data sets}. In all of the experiments, the data points are not splitted into the training and test data and all of them are considered as the training data. Using these data points, different clustering algorithms are performed and the results are obtained.

In clustering algorithms, the "clustering accuracy" is applied as a criterion to evaluate the results. First, to compute the predicted label, the most frequent class label of each cluster is assigned to all of its data points. Then, the accuracy is computed by total number of data points with the correct predicted label dividing by the total number of all data. Formally it is as follows:
\begin{equation}
	ACC=\dfrac{\sum_{n=1}^{N}\delta(\hat{y}_{n},\textit{map}(\mathbf{y}_{n}))}{N}\times100,
\end{equation}
where $N$ is the number of data points, $\hat{y}_{n}$ is the real correct label, \textit{map} is a function, which computes the predicted label $\mathbf{y}_{n}$ and delta function
$\delta(s,t)=1$ when $s = t$ , otherwise it is 0 \cite{moslehi2018discriminative}. 

"Normalized mutual information (NMI)" as another measure is computed as follows:
\begin{equation}
	NMI(\hat{Y},I)=200\times\dfrac{\sum_{\hat{y}_{i}\in\hat{Y},cl_{j}\in I}p(\hat{y}_{i},cl_{j})log\dfrac{p(\hat{y}_{i},cl_{j})}{p(\hat{y}_{i})p(cl_{j})}}{H(\hat{Y})+H(I)},
\end{equation}
where $\hat{Y}$ and $I$ are the sets of true labels and cluster indicators. $p(\hat{y}_{i})$, $p(cl_{j})$, and $p(\hat{y}_{i},cl_{j})$ are the probabilities that each randomly selected data point might belong to the class $\hat{y}_{i}$, cluster $cl_{j}$, and in the intersection of $\hat{y}_{i}$ and $cl_{j}$. The functions $H(\hat{Y})$ and $H(I)$ are the entropy of $\hat{Y}$ and $I$, respectively \cite{moslehi2018discriminative}.

In all of these experiments, the number of clusters is set to be equivalent to the number of classes. Thus, for our proposed method SGP-MIC, the parameter $M$ is set to the number of classes, and each component of this model indicates a unified cluster. In this method, each data point $\mathbf{y}_{n}$ belongs to the $m$-th cluster with maximum value of $q\left(s_{n,m}\right)$. For data visualization we use $\mathbf{X}\in\Re^{N\times 2}$, to illustrate the 2 dimensional embedded data points. The stopping criteria is also reaching the predefined number of iterations. For all selected data sets except Coil20, CMC and Segment, the number of inducing variables, $N'$, is set to 50 and for these three larger data sets, it is set to 100. The low dimensional space $\mathbf{X}$ is initialized by ISOMAP algorithm. $\mathbf{X}_{u}$ is randomly selected from $\mathbf{X}$, each element of $M$ dimensional vector $\boldsymbol{\pi}$ is set to 1/$M$, and the cluster centers $\bar{\mathbf{x}}_{m}$ are set by FCM clustering algorithm. For comparison results, we applied linear kernel,
\begin{equation}
	\begin{aligned}[b]
		{k}^{\left(m\right)}(\mathbf{z},\mathbf{z'})= {\theta}^{\left(m\right)}_{\textit{ \emph{lin}}}\mathbf{z}^T\mathbf{z'}+{\theta}^{\left(m\right)}_{\textit{ \emph{bias}}}+{\theta}^{\left(m\right)}_{\textit{ \emph{white}}},
	\end{aligned}
\end{equation}
and radial basis function (RBF) kernel,
\begin{equation}
	\begin{aligned}[b]
		&{k}^{\left(m\right)}(\mathbf{z},\mathbf{z'})= {\theta}^{\left(m\right)}_{\textit{ \emph{rbf}}}\textit{\emph{exp}}\left(\frac{\gamma^{\left(m\right)}}{2}(\mathbf{z}-\mathbf{z'})^T(\mathbf{z}-\mathbf{z'})\right)\\
		&\; \; \; \; \;\; \; \; \; \;\; \; \; \; \;\; \; \; \; \; \; \; \;+{\theta}^{\left(m\right)}_{\textit{ \emph{bias}}}+{\theta}^{\left(m\right)}_{\textit{ \emph{white}}},
	\end{aligned}
\end{equation}
where ${k}^{\left(m\right)}(\mathbf{z},\mathbf{z'})$ is an element of the kernel matrix $\mathbf{K}^{\left(m\right)}$ and ${\theta}^{\left(m\right)}_{\textit{ \emph{lin}}}$, ${\theta}^{\left(m\right)}_{\textit{ \emph{rbf}}}$, ${\theta}^{\left(m\right)}_{\textit{ \emph{bias}}}$, ${\theta}^{\left(m\right)}_{\textit{ \emph{white}}}$ and $\gamma^{\left(m\right)}$ are kernel parameters associated with the $m$-th component of the mixture.
The kernel parameters are initialized as ${\theta}^{\left(m\right)}_{\textit{ \emph{lin}}}={\theta}^{\left(m\right)}_{\textit{ \emph{rbf}}}=\gamma^{\left(m\right)}=1$ and ${\theta}^{\left(m\right)}_{\textit{ \emph{bias}}}={\theta}^{\left(m\right)}_{\textit{ \emph{white}}}=\textit{\emph{exp}}(-2)$ and then they are updated during the optimization. These parameters are shared between $\mathbf{K}_{uu}^{\left(m\right)}$, $\mathbf{K}_{fu}^{\left(m\right)}$ and $\mathbf{K}_{uf}^{\left(m\right)}$. For computing $\mathbf{K}_{uu}^{\left(m\right)}$, $\mathbf{z}$ and $\mathbf{z'}$ are chosen from $\mathbf{X}_{u}$ and for computing $\mathbf{K}_{fu}$, $\mathbf{z}$, and $\mathbf{z'}$ are selected from $\mathbf{X}$ and $\mathbf{X}_{u}$, respectively. The value of $\beta$ is set as a function of data's variance, i.e. ${\beta= (\dfrac{1}{0.5\sqrt{mean(var(Y))}})^{2}}$.

\begin{figure*}[b]
	\centering
	\setlength\tabcolsep{1.5pt}	
	\begin{tabular}{ c  c  c}
		
		\includegraphics[scale=0.27]{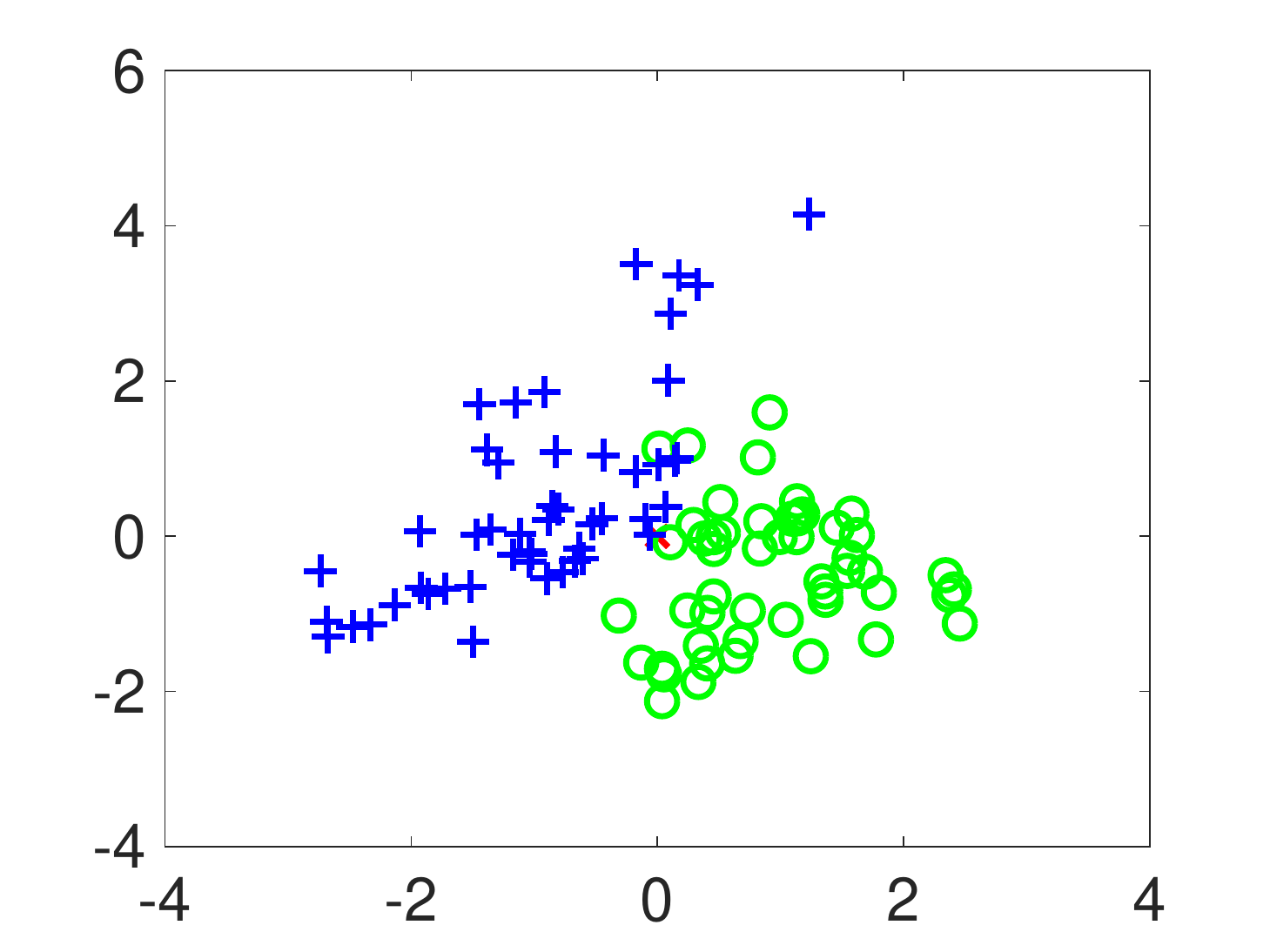} &\includegraphics[scale=0.27]{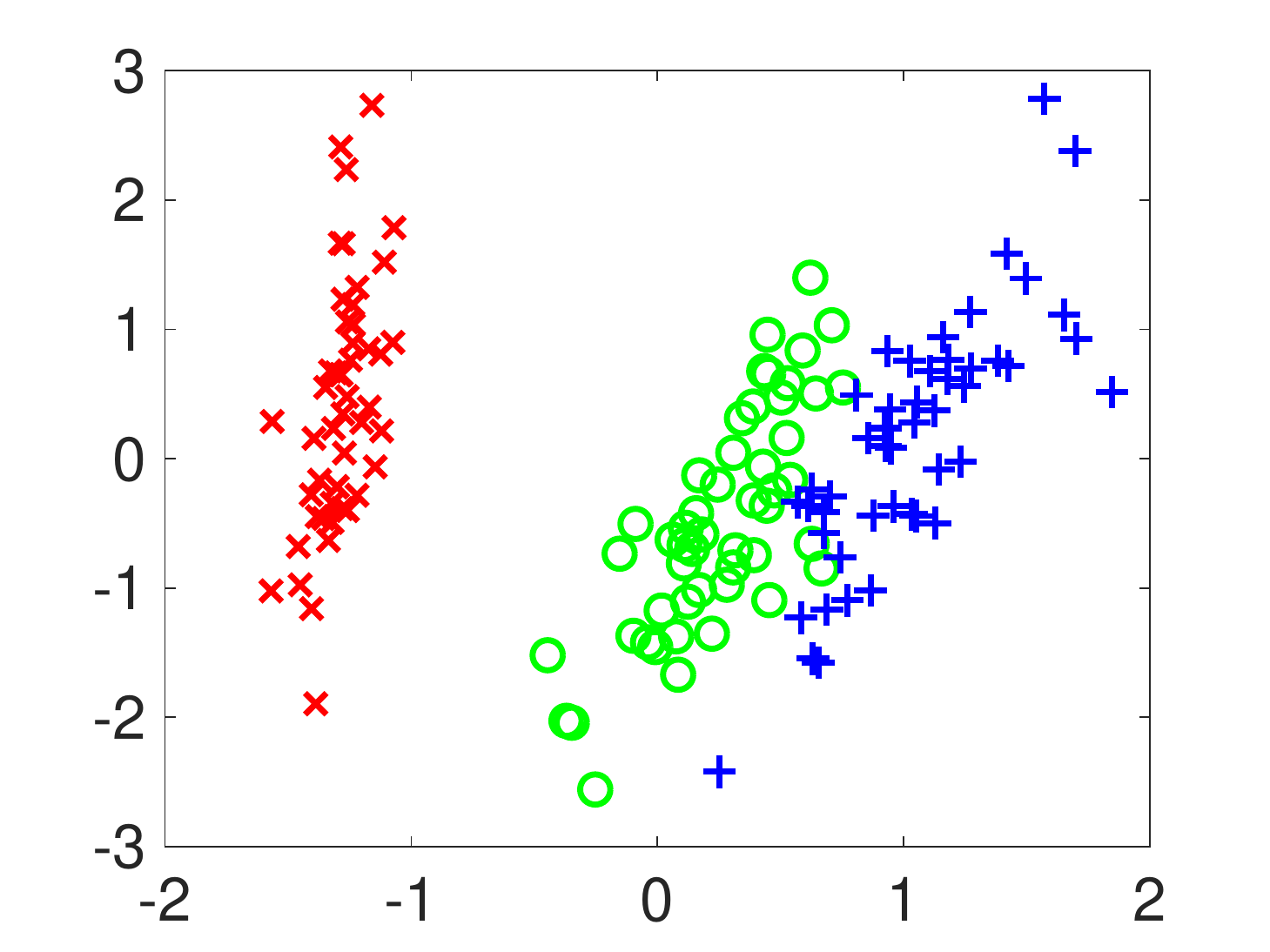}&\includegraphics[scale=0.27]{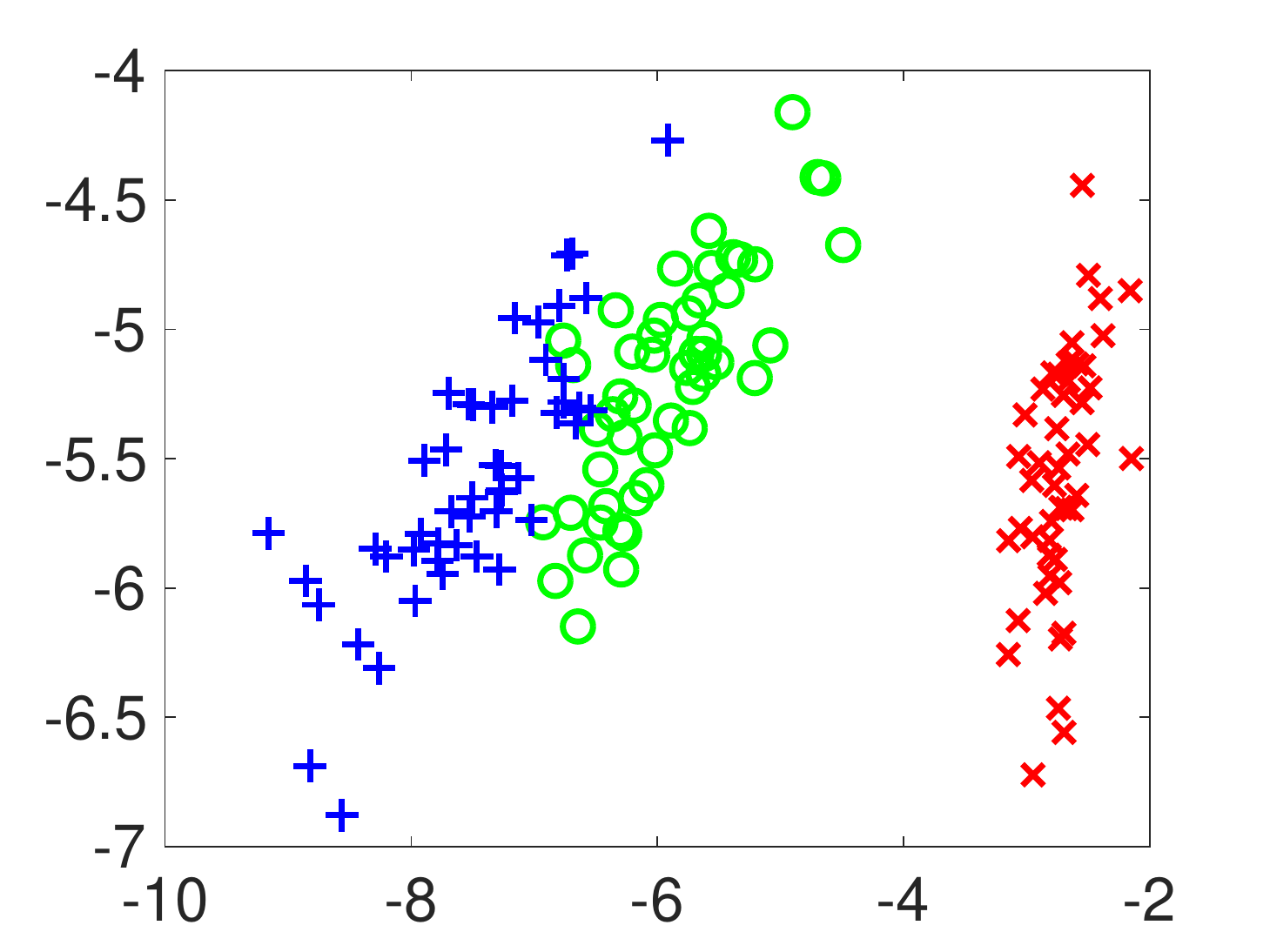} \\
		(a) ISOMAP & (b) PPCA & (c) AML \\
		ACC=76.67 & ACC=76.00 & ACC=77.33 \\
		\includegraphics[scale=0.27]{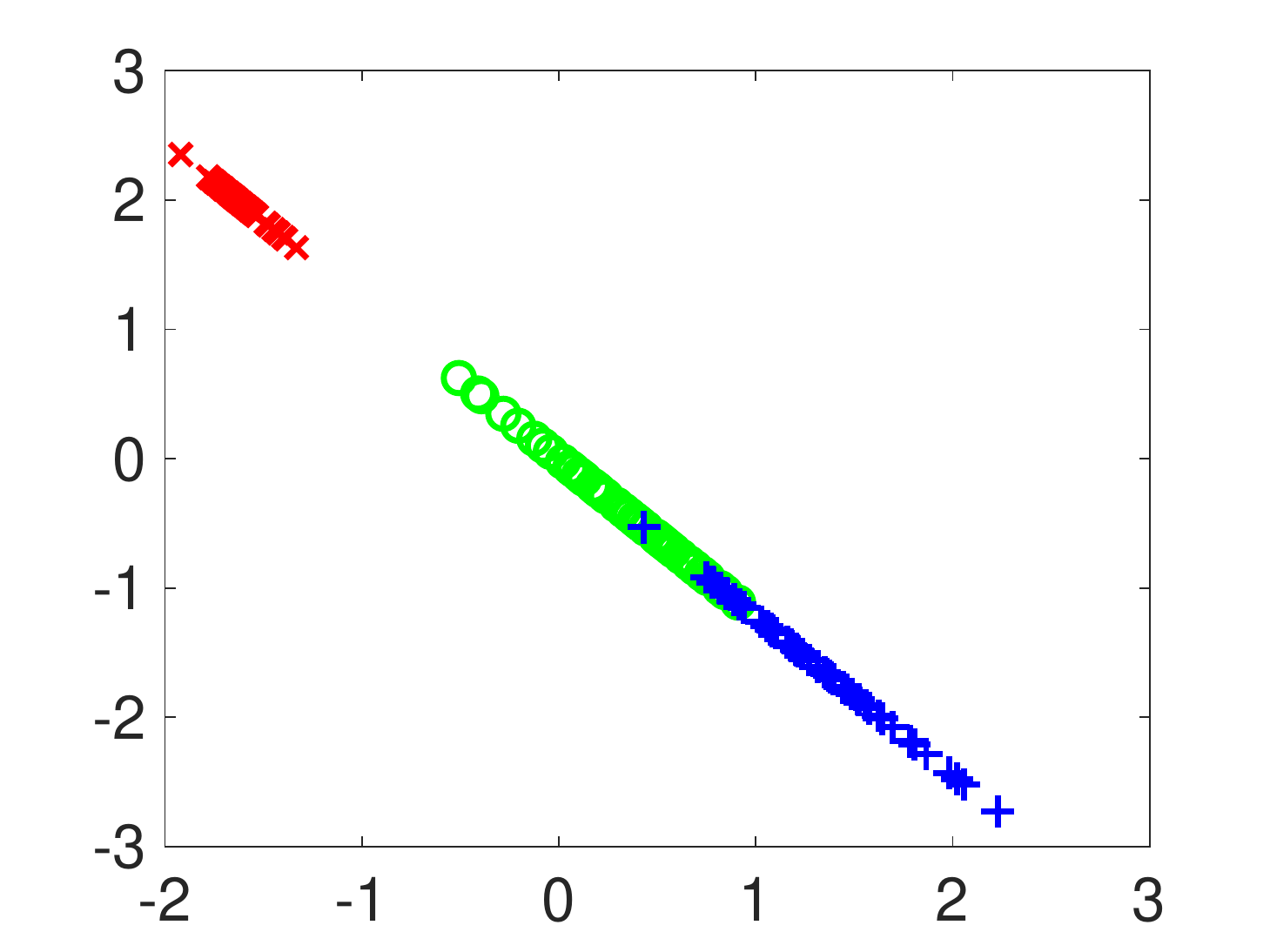}& \includegraphics[scale=0.27]{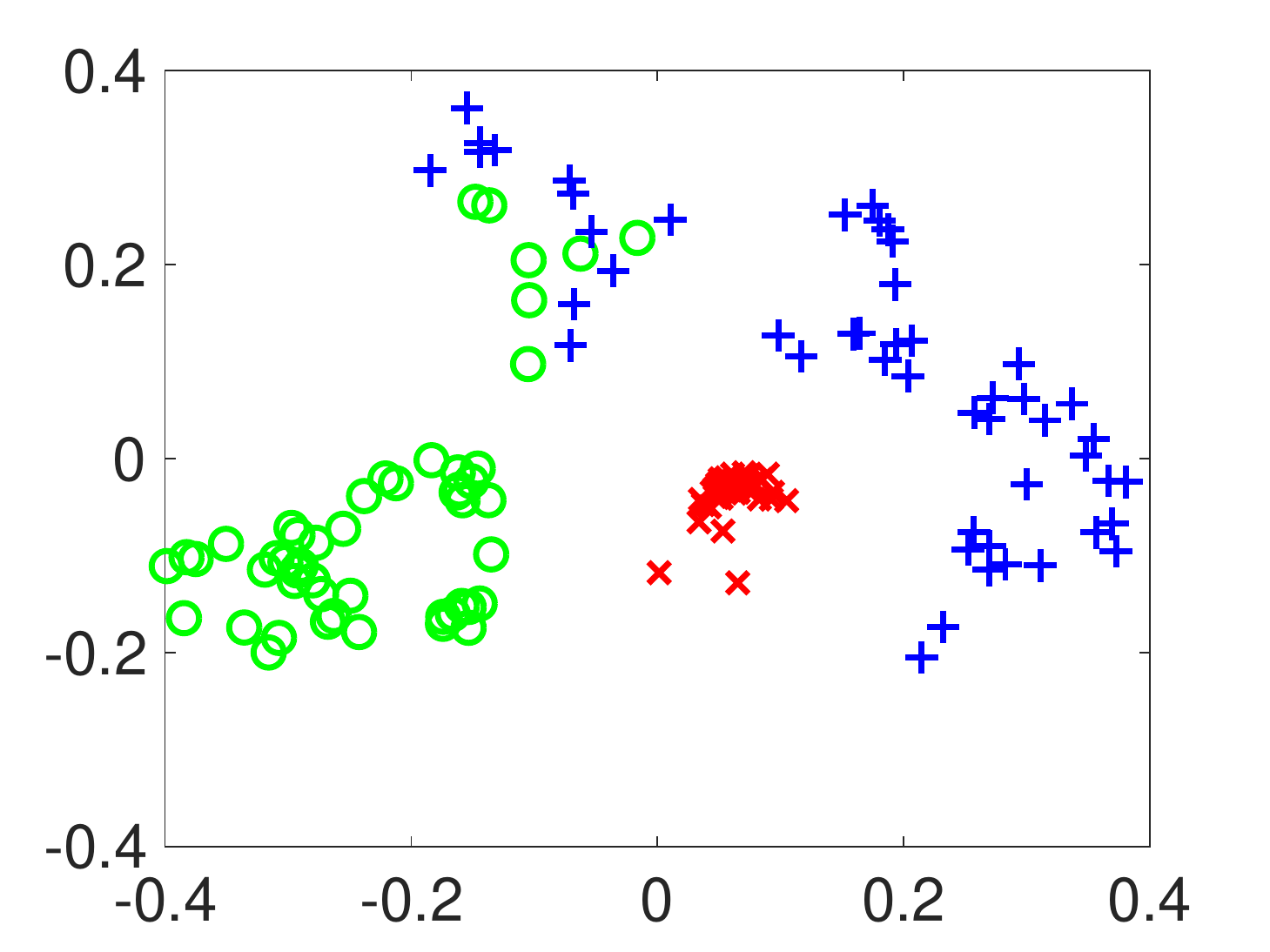} &\includegraphics[scale=0.27]{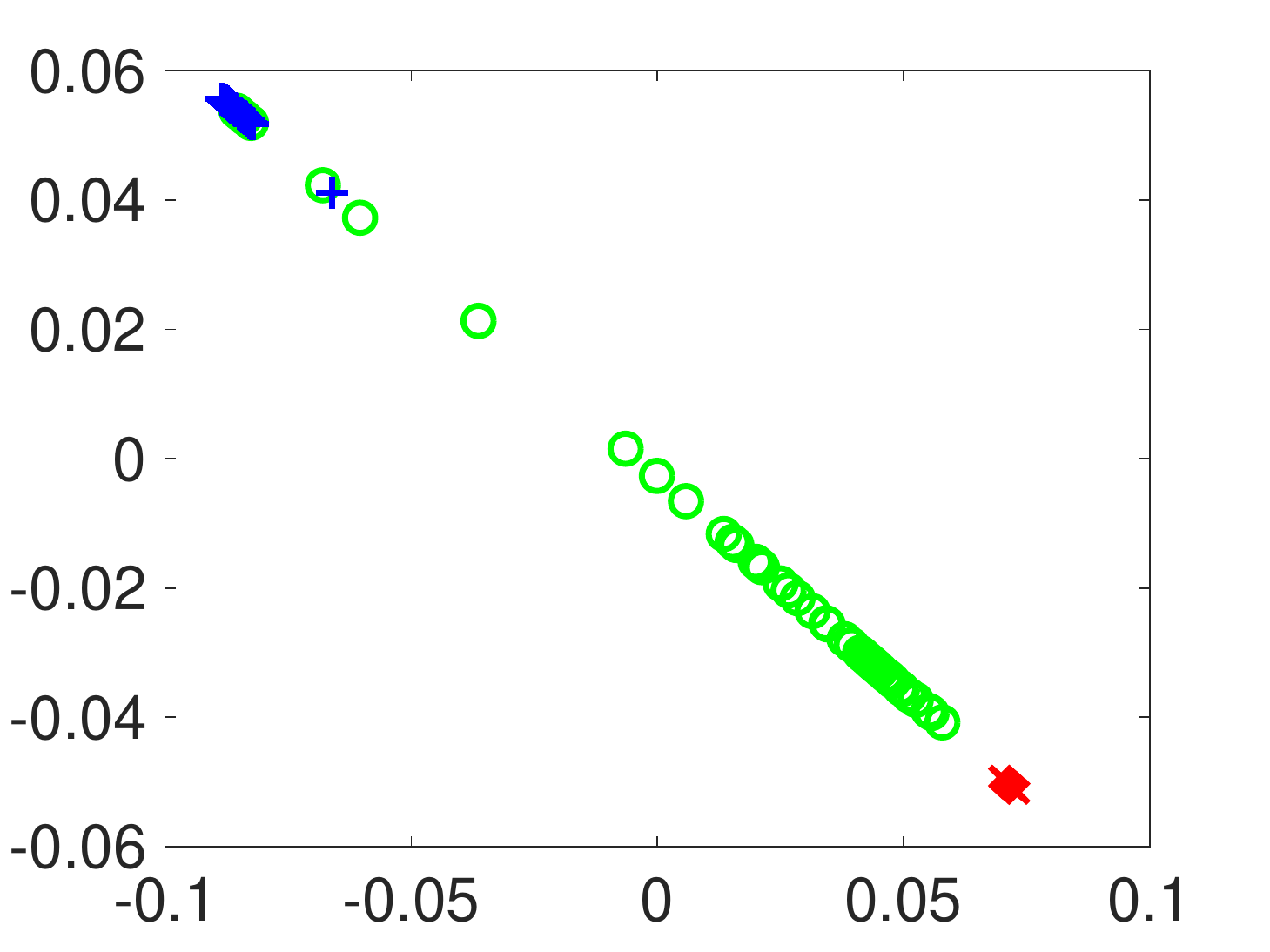}\\ 
		
     	(d) UNCA &(e) GP-LVM & (f) SGP-MIC \\ 			 
		
		ACC=88.27 & ACC=74.00 & ACC=93.33 \\ 			 
	\end{tabular}
	\caption{2D visualization of the Iris data by using (a) ISOMAP, (b) PPCA, (c) AML, (d) UNCA, (e) GP-LVM, and (f) SGP-MIC methods.}\footnotesize
	\label{figure:twoDVisulisationIris}
\end{figure*}
\begin{figure*}[t]
	\centering
	\setlength\tabcolsep{1.5pt}			
	\begin{tabular}{c c c }			
		\includegraphics[scale=0.27]{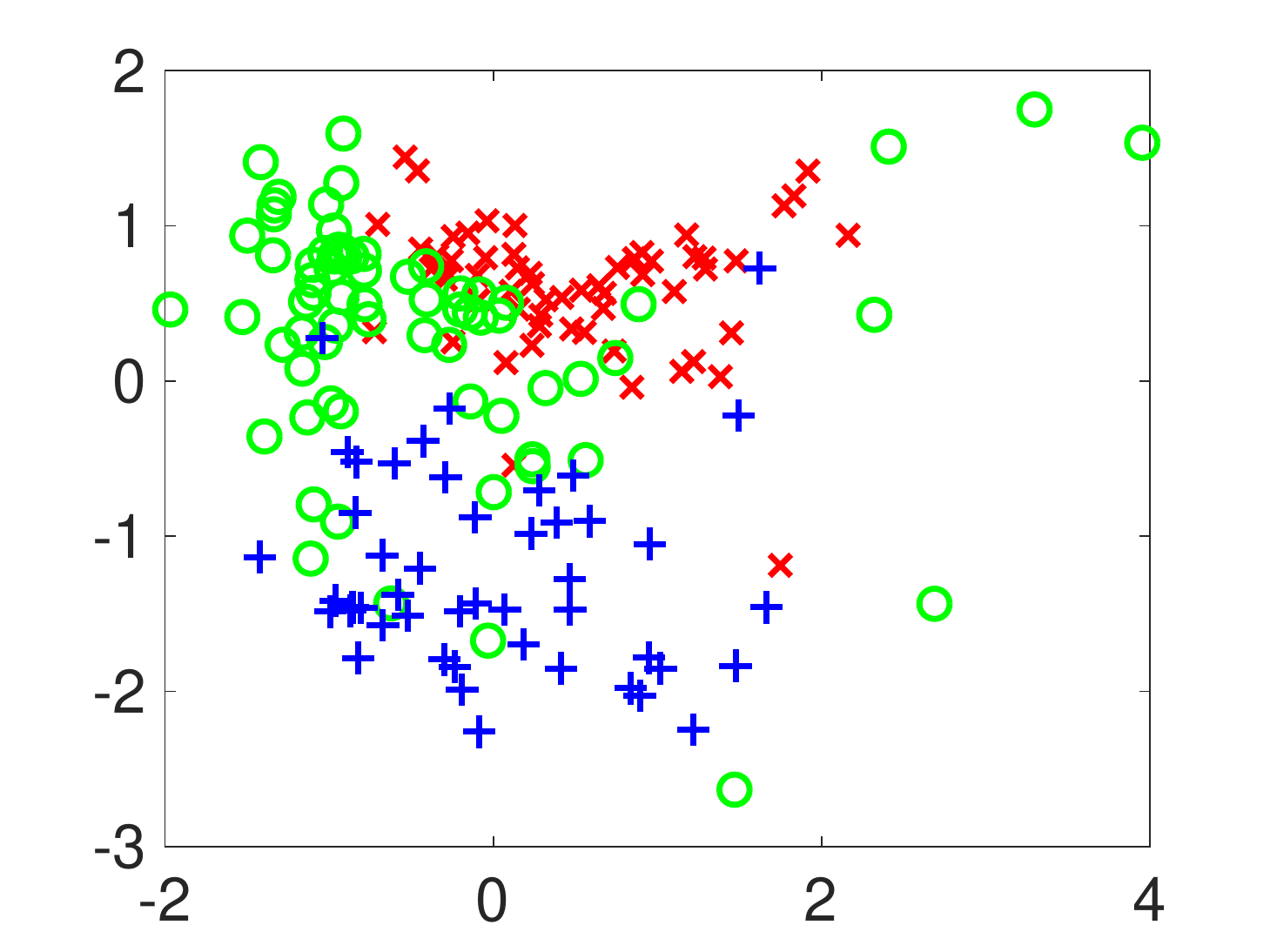} &\includegraphics[scale=0.27]{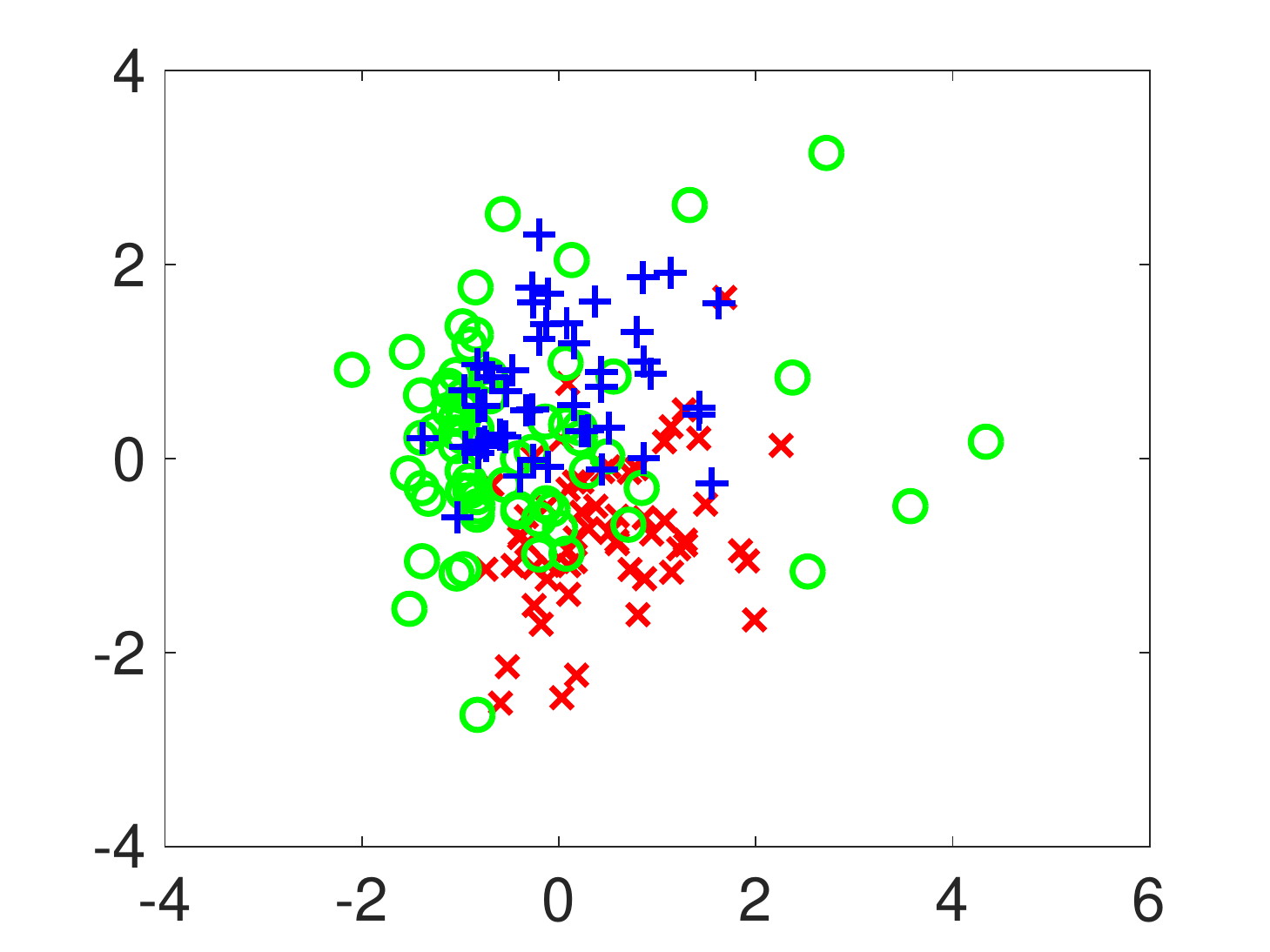}  & \includegraphics[scale=0.27]{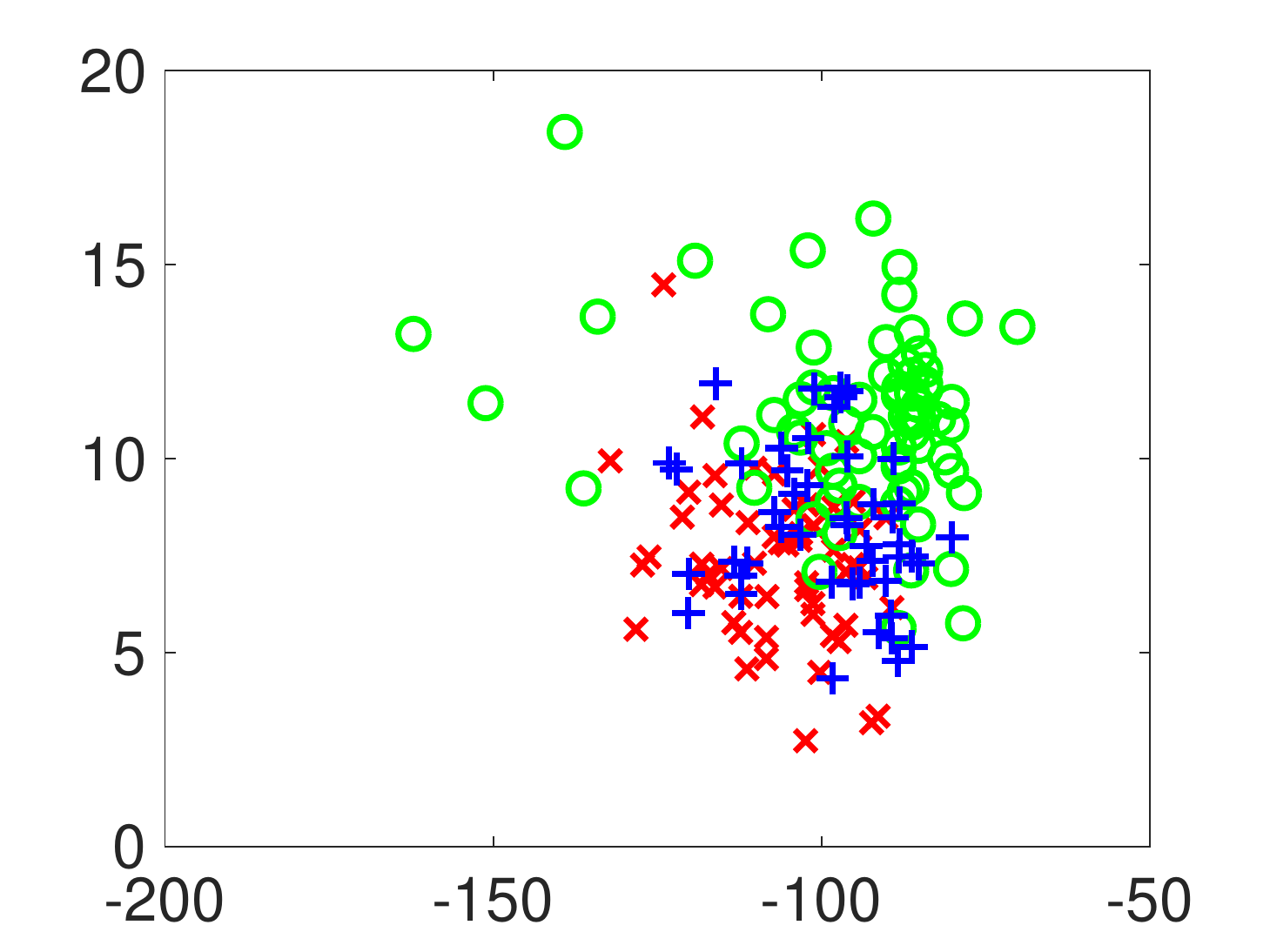}\\
		(a) ISOPMAP&(b) PPCA & (c) AML \\
		ACC=72.47 & ACC=61.80 & ACC=60.67\\ \includegraphics[scale=0.27]{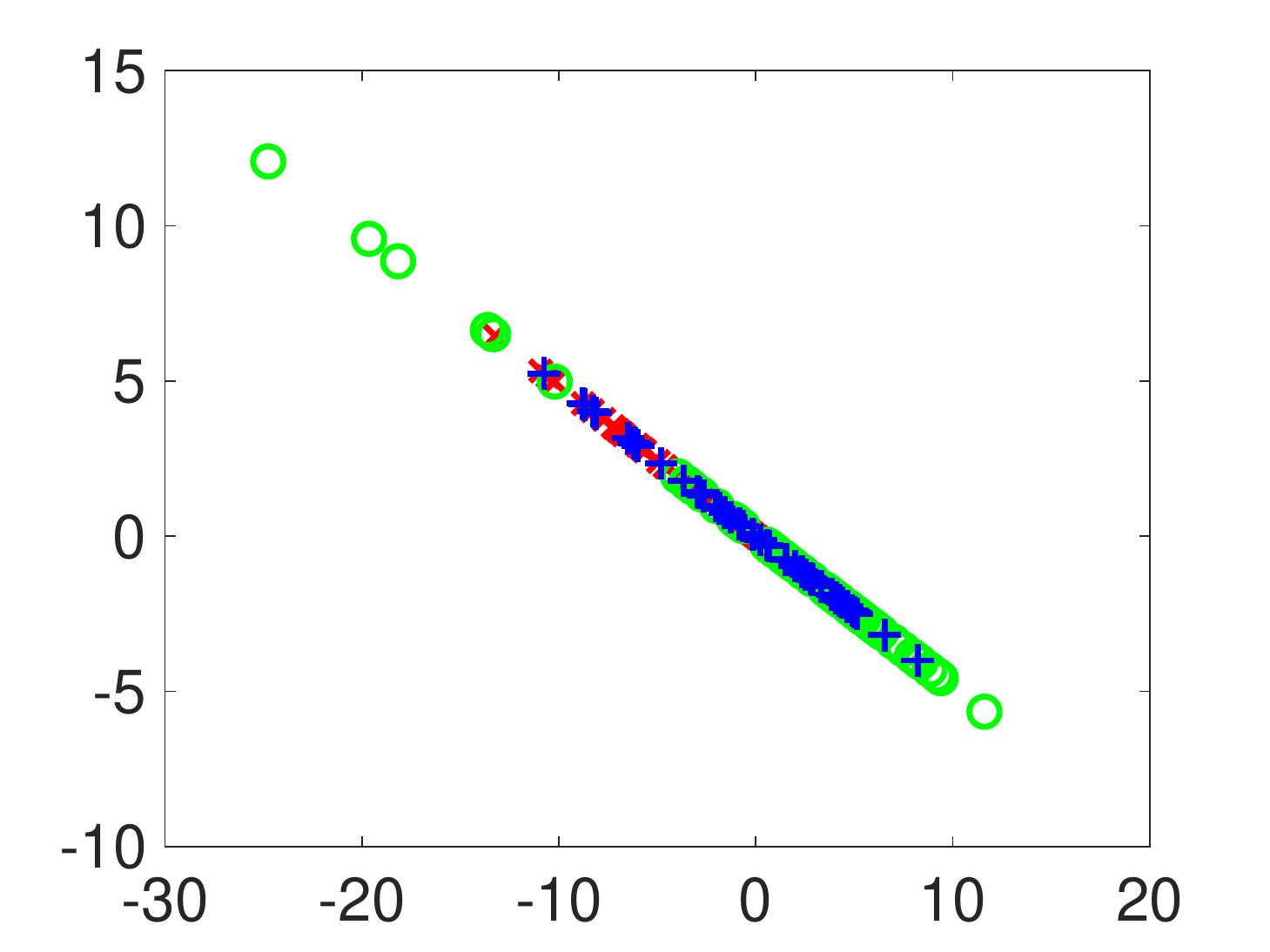} & \includegraphics[scale=0.27]{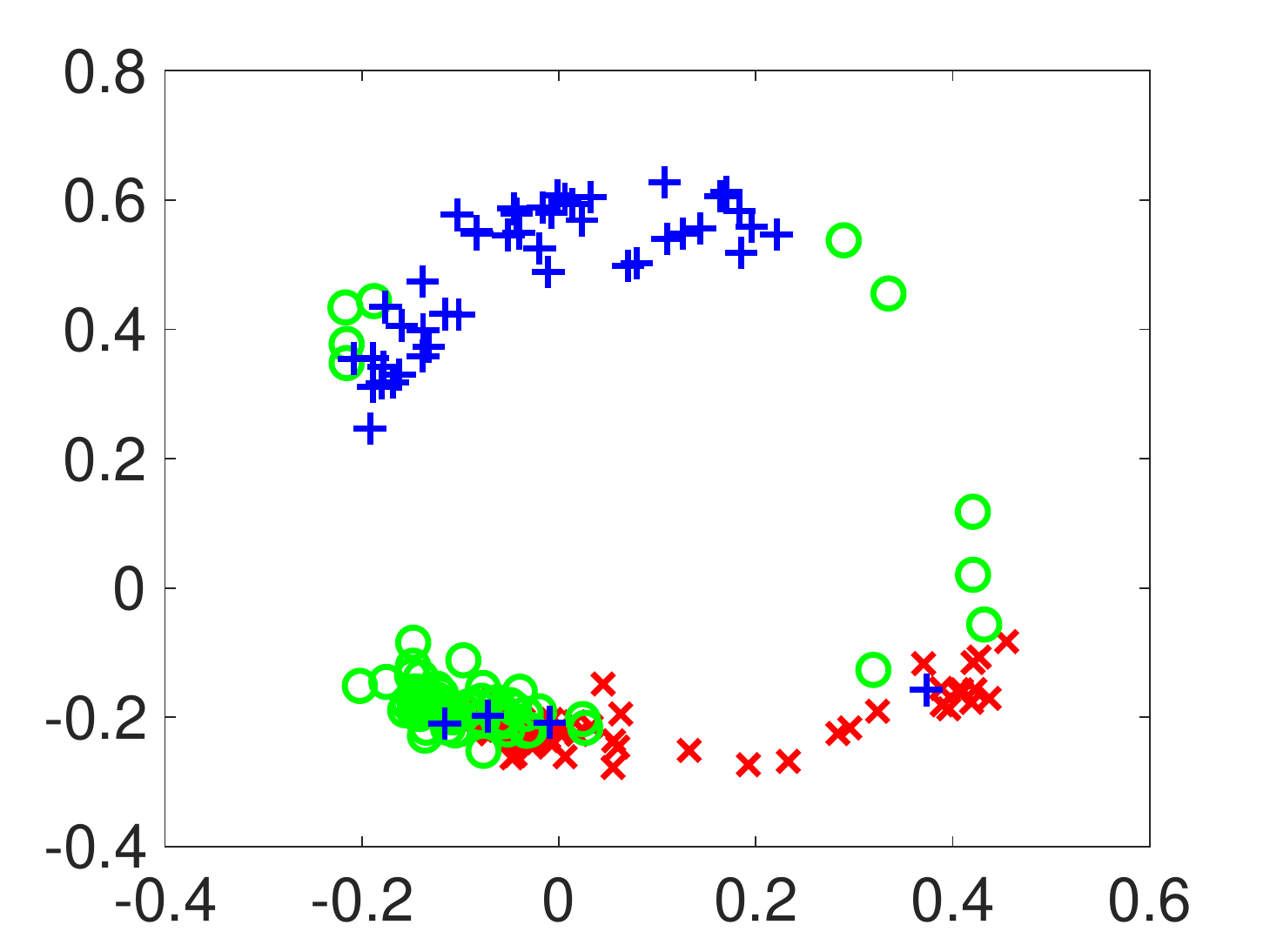} &\includegraphics[scale=0.27]{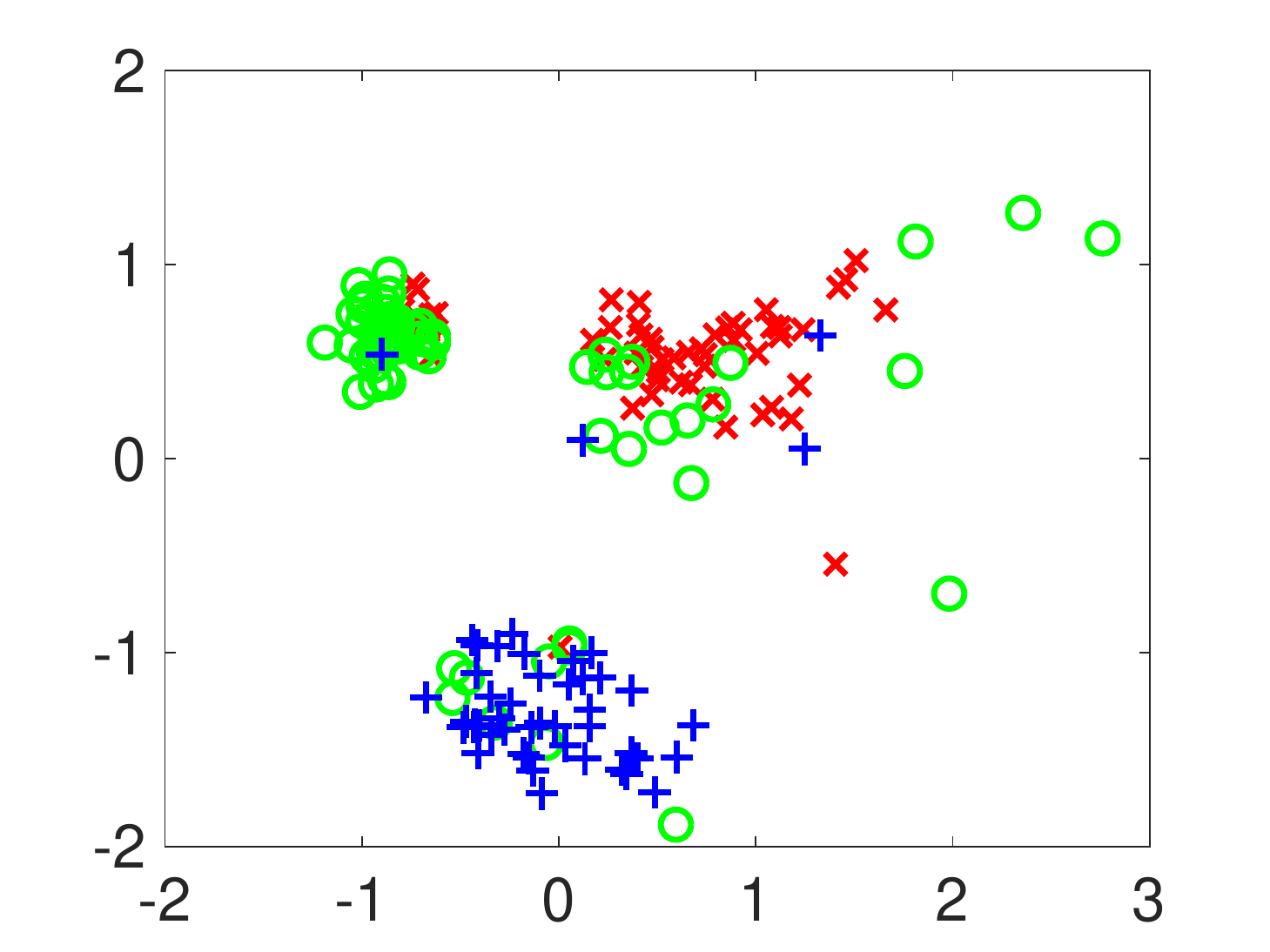} \\ 
		(d) UNCA &(e) GP-LVM & (f) SGP-MIC \\ 	 	 
		 ACC=50.45 & ACC=69.10 &ACC=75.28 \\ 	 	 
		
	\end{tabular}
	\caption{2D visualization of the Wine data by using (a) ISOMAP, (b) PPCA, (c) AML, (d) UNCA, (e) GP-LVM, and (f) SGP-MIC methods.}\footnotesize
	\label{figure:twoDVisulisationWine}
\end{figure*}

\begin{figure*}[t]
	\centering
	\setlength\tabcolsep{1.5pt}			
	\begin{tabular}{c c }			
		\includegraphics[scale=0.75]{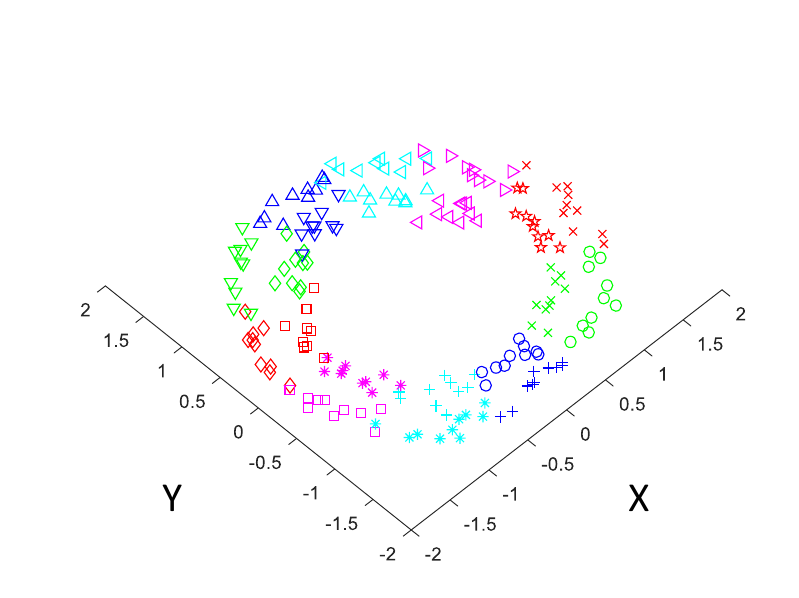} &\includegraphics[scale=0.75]{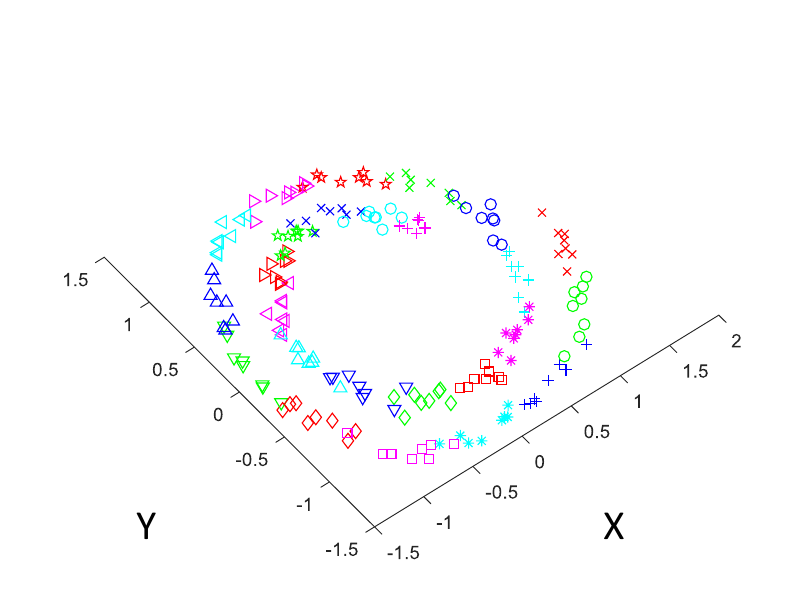} \\
		(a) Input data&(b)  SGP-MIC  \\
		\includegraphics[scale=0.75]{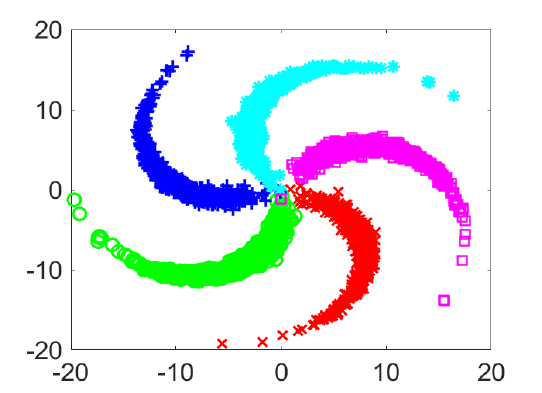} &\includegraphics[scale=0.75]{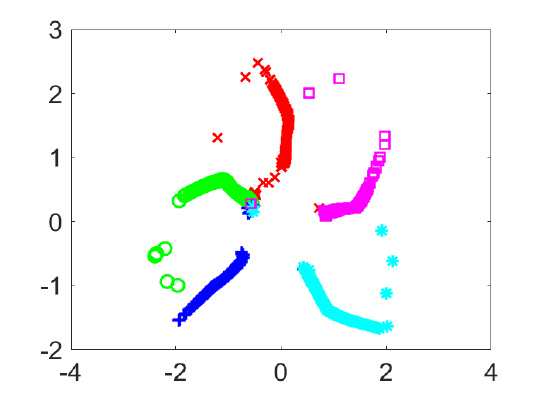} \\
		(a) Input data&(b)  SGP-MIC  \\\\  	 
		
	\end{tabular}
	\caption{2D visualization of the two spherical data sets, (a) Input data, (b) SGP-MIC.}\footnotesize
	\label{figure:twoDVisulisationSpiral}
\end{figure*}

\subsection{2D Visualization\label{sec:twoD Visualization}}

2D visualization of a data set is very natural for humans to analyze its data structure.

\textbf{Iris Data} Our first illustration of the model is in the context of a widely used benchmark
data set: we considered the Iris data set and modeling its data points into a two dimensional space after applying ISOMAP \cite{tenenbaum2000global}, PPCA \cite{tipping1999probabilistic}, AML \cite{ye2007adaptive}, UNCA \cite{qin2015unsupervised}, RBF kernel based GP-LVM \cite{lawrence2004gaussian} and SGP-MIC algorithms. The results are shown in Figure \ref{figure:twoDVisulisationIris}. The same symbols define the same class data. It is observed that after applying the SGP-MIC algorithm, the separation between the different classes of data is good, and the components appear well coordinated. The clustering accuracies of the latent space for different methods are also presented in this figure. After applying ISPOMAP, PPCA and GP-LVM we cluster data using FCM clustering algorithm. The other algorithms automatically cluster data by learning the cluster indicator matrix. In SGP-MIC, each data point $\mathbf{x}_{n}$ belongs to the $m$-th cluster with maximum value of $q\left(s_{n,m}\right)$. By SGP-MIC method, all cross points define a unified cluster, but through GP-LVM followed by FCM, some of different class data points might belong to this cluster (Figure \ref{figure:twoDVisulisationIris}(e, f)). Thus, the accuracies of SGP-MIC and GP-LVM followed by FCM methods are very different. This can be happened for all of other methods. It should be noted that this figure shows the embedded space of all data points, where different symbols show different classes and they do not define clusters.

\textbf{Wine Data} We also applied this proposed algorithm and other existing methods to the Wine data set. In Figure \ref{figure:twoDVisulisationWine} we show all data points visualized in 2-D latent space after performing different algorithms. The 2-D visualization and the clustering accursacies show the superiority of our proposed method on this data set.

\textbf{Spherical Data} We also applied this proposed algorithm to some spherical data sets. For the first data set, we generate data points on a 3D spiral and for the second data set, we create data points in 2D space created from the arcs of 5 circles. In Figure \ref{figure:twoDVisulisationSpiral}, we show all raw data points and embedded data points after running SGP-MIC method visualized in 2D latent space. For the 3D spiral data set, we just set the value of third dimension to zero. In these data sets, the raw data are in the nonlinear space but considering the nonlinear property of our method, we could learn this nonlinear structure as the accuracy and NMI of SGP-MIC on second synthetic data set are 99.32 and 97.35, respectively.

\begin{figure*}[htb]
	\centering		
	\begin{tabular}{ @{}c@{}  @{}c@{}   }		\centering	
		\includegraphics[scale=0.3]{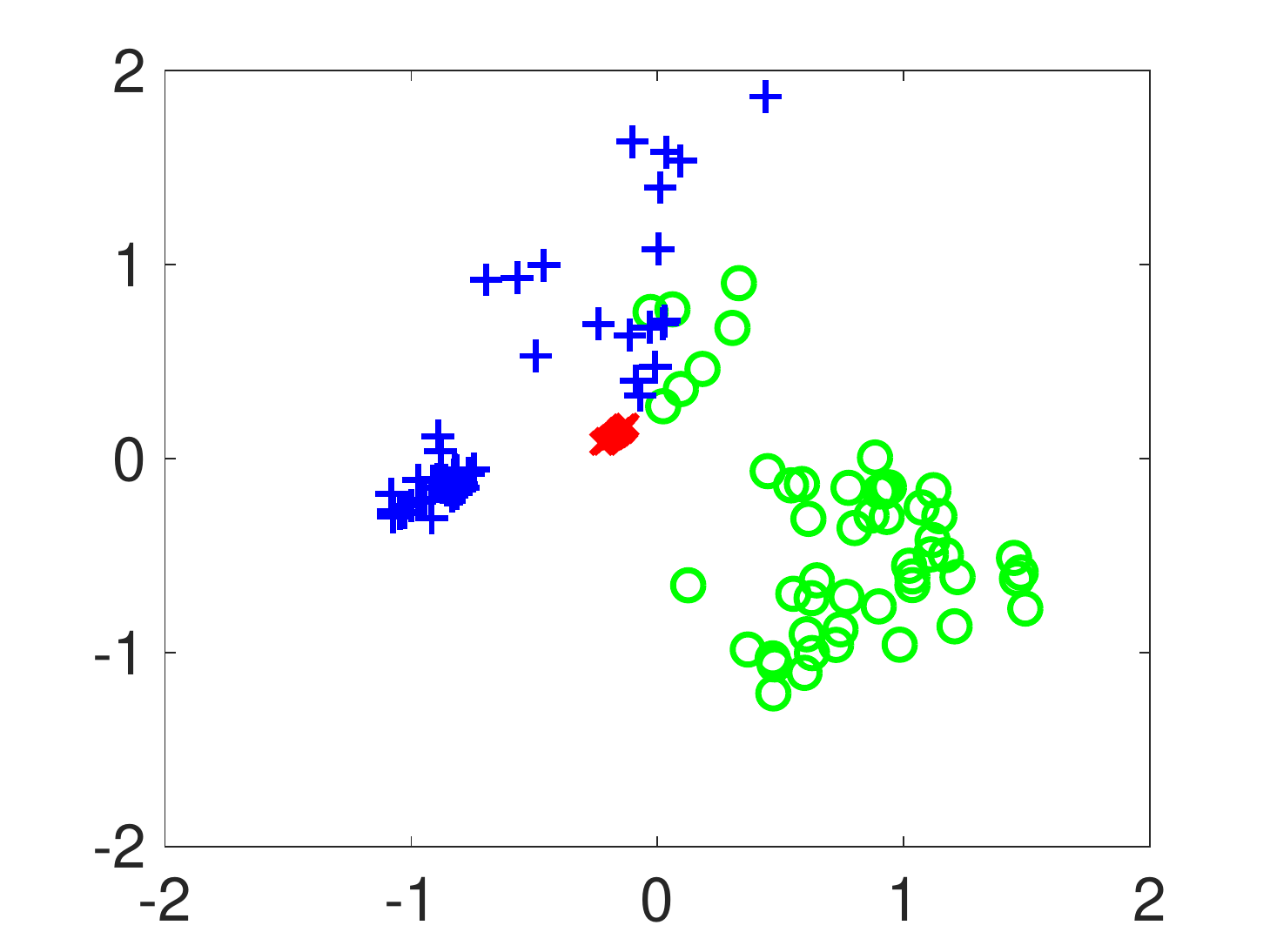} & \includegraphics[scale=0.3]{IrisRBF} \\
		(a) Iris, ACC=82.67 & (b) Iris, ACC=93.33\\ \includegraphics[scale=0.3]{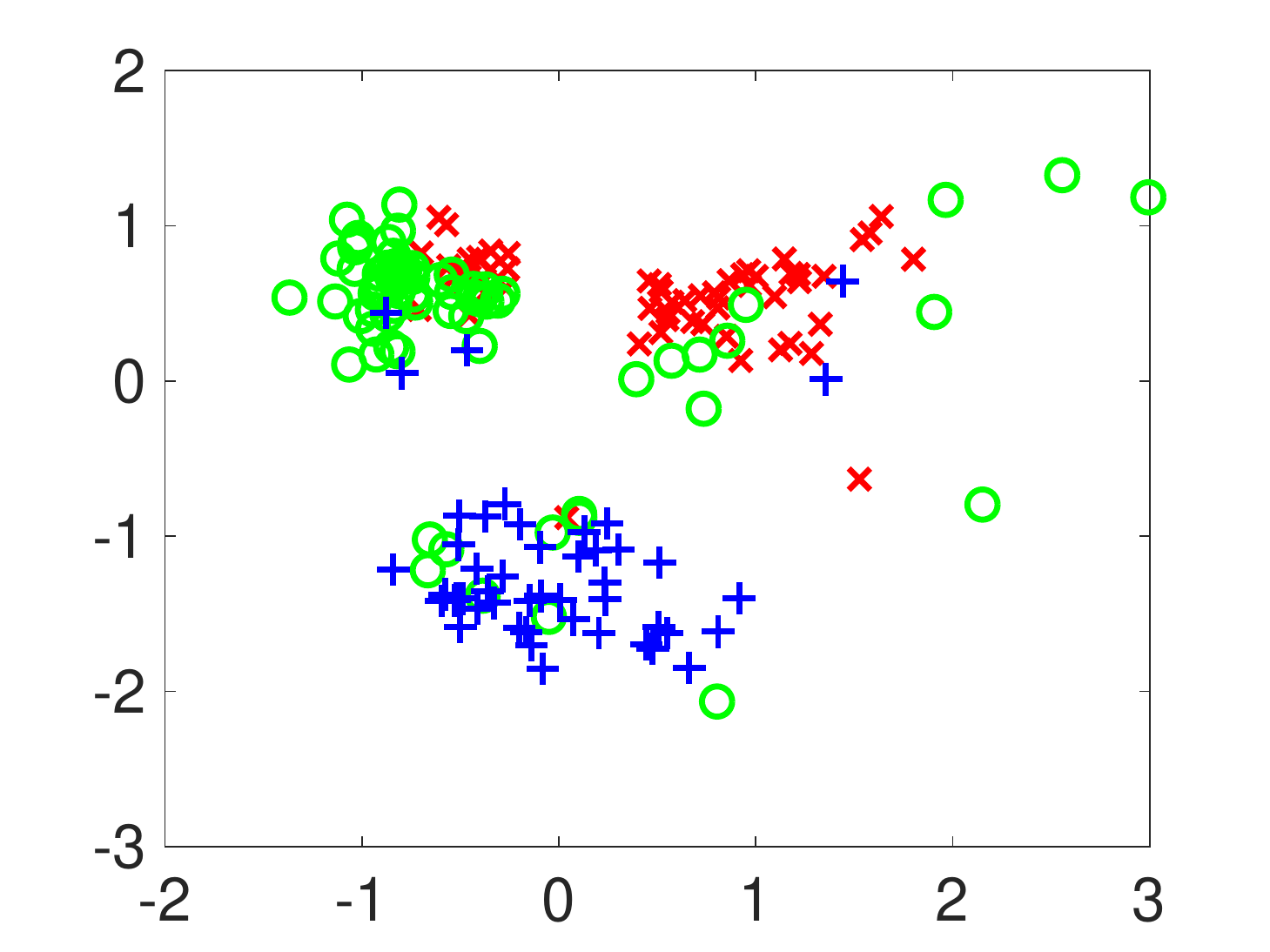}& \includegraphics[scale=0.3]{WineRBF}\\ 
		 (c) Wine, ACC=74.16 & (d) Wine, ACC=75.28 \\ 
	\end{tabular}
	\caption{(a, c) SGP-MIC with linear kernel. (b, d) SGP-MIC with RBF kernel.}\footnotesize
	\label{figure:Kernel Effect}
\end{figure*}

\subsection{Kernel Effect\label{sec:Kernel Effect}}

In this section, we are interested in evaluating our method using the linear and RBF kernels. We apply our method with these kernels on Iris and Wine data sets. 2-D visualization of these data sets and their clustering accuracies are shown in Figure \ref{figure:Kernel Effect}. The results show that RBF kernel behaves better than the linear kernel. Through the linear kernel, it may not be possible to represent the non-linear structure of data points in the low-dimensional embedded space. This is one of the main advantages of our proposed method, addresses the weakness of other approaches which rely on linear mapping.
\newcolumntype{g}{>{\columncolor{LightRed}}c}
\begin{table*}[htb]
	\centering

	\caption{ACC/NMI comparison of different methods after dimension reduction on 11 data sets.} \label{table:ACC comparison}
	\begin{adjustbox}{width=1\textwidth}
	\setlength\tabcolsep{0.40pt}
	\begin{tabular}{c c c c c c c c c c c c c c c}
		\toprule 
		\multicolumn{3}{l}{} & PCA & ISOMAP & PPCA & FGPLVM & AE & VAE & AML & UNCA & DEC & SDC & DEEPC & SGP-MIC \\
		\multicolumn{3}{l}{} & -FCM & -FCM & -FCM & -FCM & -FCM & -FCM &  &  &  &  &  &  \\
		\cmidrule{1-15}
		\multirow{22}{*}{\begin{turn}{90}Accuracy\end{turn}} & Iris & 50(1) & 91.33$\pm$0.00 & 71.33$\pm$0.00 & 91.33$\pm$0.00 & 85.67$\pm$4.87 & 85.53$\pm$0.32 & 38.13$\pm$2.13 & 91.60$\pm$0.84 & 76.40$\pm$6.87 & - & 90.67$\pm$0.00 & \fontseries{b}\selectfont{96.66$\pm$0.84} & 92.33$\pm$0.65 \\
		& & 95(2) & 88.67$\pm$0.00 & 76.67$\pm$0.00 & 76.00$\pm$0.00 & 83.00$\pm$3.67 & 87.53$\pm$1.54 & 37.93$\pm$1.65 & 77.33$\pm$0.00 & 86.20$\pm$3.82 & \fontseries{b}\selectfont{95.33$\pm$0.00} & 89.33$\pm$0.00 & 94.93$\pm$3.83 & 92.73$\pm$3.38 \\
		& & 99(3) & 89.33$\pm$0.00 & 69.33$\pm$0.00 & 76.53$\pm$0.28 & 76.27$\pm$3.07 & 89.20$\pm$1.53 & 38.93$\pm$3.30 & 76.67$\pm$0.00 & 87.73$\pm$5.67 & \fontseries{b}\selectfont{96.00$\pm$0.00} & 89.33$\pm$0.00 & 94.00$\pm$5.09 & 85.60$\pm$0.90 \\
		\cmidrule{2-15}
		& Wine & 50(1) & 55.06$\pm$0.00 & 55.06$\pm$0.00 & 55.06$\pm$0.00 & 55.51$\pm$0.24 & 69.27$\pm$0.46 & 49.33$\pm$6.84 & 54.61$\pm$0.52 & 52.24$\pm$3.35 & - & 70.22$\pm$0.00 & \fontseries{b}\selectfont{70.78$\pm$0.00} & 55.62$\pm$0.00 \\
		& & 95(2) & 55.06$\pm$0.00 & 72.47$\pm$0.00 & 61.80$\pm$0.00 & 73.15$\pm$0.36 & 69.04$\pm$0.18 & 45.51$\pm$9.75 & 60.73$\pm$2.52 & 50.71$\pm$2.66 & 57.86$\pm$1.24 & 70.22$\pm$0.00 & 70.28$\pm$0.16 & \fontseries{b}\selectfont{75.28$\pm$0.26} \\
		& & 99(4) & 55.06$\pm$0.00 & 78.65$\pm$0.00 & \fontseries{b}\selectfont{89.89$\pm$0.00} & 79.83$\pm$0.18 & 69.04$\pm$0.18 & 44.21$\pm$9.35 & 76.46$\pm$15.34 & 48.20$\pm$4.41 & 57.52$\pm$0.71 & 70.22$\pm$0.00 & 70.56$\pm$0.27 & 77.75$\pm$0.54 \\
		\cmidrule{2-15}
		& Sonar & 50(2) & 55.77$\pm$0.00 & 53.37$\pm$0.00 & 53.37$\pm$0.00 & 55.82$\pm$3.11 & 54.42$\pm$0.90 & 53.89$\pm$1.07 & 55.05$\pm$0.91 & 57.19$\pm$0.98 & \fontseries{b}\selectfont{58.17$\pm$0.00} & 55.29$\pm$0.00 & 54.90$\pm$1.96 & 53.37$\pm$0.00 \\
		& & 95(17) & 55.29$\pm$3.09 & 57.02$\pm$4.16 & 55.00$\pm$2.19 & 56.49$\pm$3.62 & 54.76$\pm$0.97 & 53.85$\pm$1.01 & 53.37$\pm$0.00 & 55.30$\pm$1.05 & 56.73$\pm$0.91 & 55.77$\pm$0.00 & 54.56$\pm$2.05 & \fontseries{b}\selectfont{57.93$\pm$3.71} \\
		& & 99(29) & 55.29$\pm$2.19 & 53.75$\pm$1.22 & 54.47$\pm$1.81 & 55.58$\pm$2.97 & 54.90$\pm$0.96 & 53.89$\pm$1.00 & 55.14$\pm$1.81 & 54.00$\pm$1.03 & 54.32$\pm$0.00 & \fontseries{b}\selectfont{56.25$\pm$0.00} & 55.14$\pm$3.24 & 54.33$\pm$1.09 \\
		\cmidrule{2-15}
		& WDBC & 50(1) & 85.41$\pm$0.00 & 85.24$\pm$0.00 & 85.41$\pm$0.00 & 85.24$\pm$0.00 & 85.34$\pm$0.15 & 71.86$\pm$9.19 & 85.24$\pm$0.00 & 64.36$\pm$0.51 & - & 85.41$\pm$0.00 & 88.40$\pm$0.34 & \fontseries{b}\selectfont{88.40$\pm$0.00} \\
		& & 95(1) & 85.41$\pm$0.00 & 85.24$\pm$0.00 & 85.41$\pm$0.00 & 85.24$\pm$0.00 & 85.34$\pm$0.15 & 71.86$\pm$9.19 & 85.24$\pm$0.00 & 64.00$\pm$0.51 & - & 85.41$\pm$0.00 & 88.40$\pm$0.34 & \fontseries{b}\selectfont{88.40$\pm$0.00} \\
		& & 99(2) & 85.41$\pm$0.00 & 86.12$\pm$0.00 & 85.41$\pm$0.00 & 85.94$\pm$0.00 & 85.24$\pm$0.17 & 67.15$\pm$8.65 & 85.06$\pm$0.00 & 64.00$\pm$0.77 & 62.74$\pm$0.00 & 85.41$\pm$0.00 & \fontseries{b}\selectfont{88.55$\pm$0.18} & 87.70$\pm$0.00 \\ 					  
		\cmidrule{2-15}
		& Aust & 50(1) & 56.09$\pm$0.00 & 56.09$\pm$0.00 & 56.09$\pm$0.00 & 57.04$\pm$2.57 & 56.09$\pm$0.00 & 55.57$\pm$0.18 & 55.94$\pm$0.00 & 55.80$\pm$0.00 & - & 56.23$\pm$0.00 & \fontseries{b}\selectfont{69.13$\pm$0.29} & 58.70$\pm$0.00 \\
		& & 95(1) & 56.09$\pm$0.00 & 56.09$\pm$0.00 & 56.09$\pm$0.00 & 57.04$\pm$2.57 & 56.09$\pm$0.00 & 55.57$\pm$0.18 & 55.94$\pm$0.00 & 55.80$\pm$0.00 & - & 56.23$\pm$0.00 & \fontseries{b}\selectfont{69.13$\pm$0.29} & 58.70$\pm$0.00 \\
		& & 99(1) & 56.09$\pm$0.00 & 56.09$\pm$0.00 & 56.09$\pm$0.00 & 57.04$\pm$2.57 & 56.09$\pm$0.00 & 55.57$\pm$0.18 & 55.94$\pm$0.00 & 55.80$\pm$0.00 & - & 56.23$\pm$0.00 & \fontseries{b}\selectfont{69.13$\pm$0.29} & 58.70$\pm$0.00 \\
		\cmidrule{2-15}
		& Segment & 50(2) & 45.95$\pm$0.52 & 44.81$\pm$1.69 & 44.82$\pm$0.13 & 45.27$\pm$1.72 & \fontseries{b}\selectfont{54.95$\pm$7.68} & 19.61$\pm$4.42 & 47.01$\pm$2.30 & 29.23$\pm$2.82 & 50.14$\pm$1.32 & 47.66$\pm$0.00 & 41.87$\pm$7.04 & 46.42$\pm$1.65 \\
		& & 95(4) & 56.06$\pm$2.20 & 65.45$\pm$1.47 & 57.32$\pm$0.00 & \fontseries{b}\selectfont{65.36$\pm$1.39} & 52.00$\pm$6.66 & 17.42$\pm$5.63 & 53.03$\pm$3.10 & 31.36$\pm$2.28 & 50.74$\pm$0.91 & 54.81$\pm$0.00 & 50.32$\pm$7.75 & 65.28$\pm$0.63  \\
		& & 99(6) & 58.87$\pm$2.51 & 46.10$\pm$0.81 & 65.36$\pm$0.00 & 46.19$\pm$0.75 & 51.33$\pm$5.76 & 18.31$\pm$4.29 & 62.10$\pm$7.24 & 30.82$\pm$1.72 & 50.62$\pm$0.88 & 57.19$\pm$0.00 & 48.29$\pm$6.88 & \fontseries{b}\selectfont{70.97$\pm$1.42} \\
		\cmidrule{2-15}
		& BC & 50(1) & 95.75$\pm$0.00 & 95.46$\pm$0.00 & 95.75$\pm$0.00 & 95.14$\pm$0.52 & 95.68$\pm$0.16 & 65.01$\pm$0.00 & 96.19$\pm$0.00 & 86.61$\pm$5.02 & - & 96.19$\pm$0.00 & 96.57$\pm$0.11 & \fontseries{b}\selectfont{97.12$\pm$0.14} \\
		& & 95(7) & 95.61$\pm$0.00 & 95.02$\pm$0.00 & \fontseries{b}\selectfont{97.51$\pm$0.00} & 95.72$\pm$0.36 & 95.89$\pm$0.24 & 65.01$\pm$0.00 & 94.73$\pm$0.00 & 91.43$\pm$2.73 & 96.19$\pm$0.00 & 96.05$\pm$0.00 & 96.74$\pm$0.31 & 90.11$\pm$9.55  \\
		& & 99(9) & 95.61$\pm$0.00 & 89.59$\pm$4.67 & 95.75$\pm$0.00 & 85.77$\pm$11.55 & 95.77$\pm$0.16 & 65.01$\pm$0.00 & 95.17$\pm$0.00 & 88.44$\pm$4.32 & 96.04$\pm$0.00 & 96.05$\pm$0.00 & 96.58$\pm$1.66 & \fontseries{b}\selectfont{96.66$\pm$0.12} \\
		\cmidrule{2-15}
		& CMC & 50(1) & 45.55$\pm$0.00 & 44.60$\pm$0.00 & 45.55$\pm$0.00 & 44.87$\pm$0.00 & 43.03$\pm$0.54 & 42.72$\pm$0.00 & 45.34$\pm$0.36 & 43.64$\pm$0.00 & - & 44.74$\pm$0.00 & 40.78$\pm$1.06 & \fontseries{b}\selectfont{45.91$\pm$0.19} \\
		& & 95(2) & \fontseries{b}\selectfont{45.55$\pm$0.00} & 44.40$\pm$0.00 & 42.70$\pm$0.00 & 44.30$\pm$0.00 & 44.53$\pm$1.60 & 42.72$\pm$0.00 & 44.05$\pm$0.66 & 43.47$\pm$0.14 & 44.84$\pm$0.24 & 44.67$\pm$0.00 & 39.16$\pm$1.66 & 43.69$\pm$1.12  \\
		& & 99(5) & \fontseries{b}\selectfont{45.49$\pm$0.00} & 42.70$\pm$0.00 & 42.70$\pm$0.00 & 42.70$\pm$0.00 & 44.64$\pm$1.36 & 42.71$\pm$0.00 & 42.89$\pm$0.00 & 43.71$\pm$0.17 & 44.41$\pm$0.31 & 44.81$\pm$0.00 & 38.98$\pm$1.34 & 45.11$\pm$0.82 \\        				    				  
		\cmidrule{2-15}
		& Yale & 50(3) & 53.09$\pm$1.08 & 52.97$\pm$2.29 & 53.58$\pm$1.15 & 59.82$\pm$3.69 & \fontseries{b}\selectfont{62.97$\pm$2.92} & 54.61$\pm$4.48 & 52.00$\pm$2.32 & 55.41$\pm$2.10 & 46.12$\pm$2.41 & 61.21$\pm$0.00 & 40.00$\pm$0.12 & 54.65$\pm$1.40 \\                  		  
		\toprule    \toprule 
		\multirow{22}{*}{\begin{turn}{90}NMI\end{turn}} & Iris & 50(1) & 79.41$\pm$0.00 & 44.69$\pm$0.00 & 79.41$\pm$0.00 & 66.01$\pm$9.45 & 65.80$\pm$0.76 & 0.94$\pm$0.82 &  79.41$\pm$0.00 & 43.11$\pm$27.48 & - & 78.57$\pm$0.00 & \fontseries{b}\selectfont{88.56$\pm$2.08} & 79.59$\pm$1.42 \\
		& & 95(2) & 74.19$\pm$0.00 & 53.69$\pm$0.00 & 57.81$\pm$0.00 & 61.60$\pm$5.12 & 70.68$\pm$4.21 & 0.89$\pm$0.53 & 58.73$\pm$0.00 & 69.02$\pm$12.77 & 84.97$\pm$0.00 & 75.82$\pm$0.00 & \fontseries{b}\selectfont{86.28$\pm$8.18} & 83.59$\pm$4.79 \\
		& & 99(3) & 74.96$\pm$0.00 & 41.61$\pm$2.38 & 60.47$\pm$0.17 & 51.46$\pm$5.20 & 74.94$\pm$3.63 & 1.80$\pm$1.82 & 58.25$\pm$0.00 & 76.93$\pm$7.06 & \fontseries{b}\selectfont{86.41$\pm$0.00} & 75.82$\pm$0.00 & 84.54$\pm$10.39 & 73.59$\pm$1.54 \\
		\cmidrule{2-15}
		& Wine & 50(1) & 13.42$\pm$0.00 & 14.46$\pm$0.00 & 13.42$\pm$0.00 & 14.31$\pm$0.47 & 41.78$\pm$0.27 & 11.98$\pm$9.02 & 10.87$\pm$0.76 & 11.61$\pm$3.05 & - & \fontseries{b}\selectfont{42.88$\pm$0.00} & 42.41$\pm$0.59 & 16.72$\pm$0.00 \\
		& & 95(2) & 13.42$\pm$0.00 & 37.42$\pm$0.00 & 24.93$\pm$0.74 & 37.96$\pm$0.30 & 41.82$\pm$0.22 & 8.48$\pm$14.55 & 23.67$\pm$3.05 & 9.31$\pm$5.65 & 19.63$\pm$0.00 & 42.88$\pm$0.00 & \fontseries{b}\selectfont{42.90$\pm$0.08} & 41.00$\pm$0.34 \\
		& & 99(4) & 13.42$\pm$0.00 & 49.28$\pm$0.00 & \fontseries{b}\selectfont{69.75$\pm$0.00} & 51.51$\pm$0.56 & 41.82$\pm$0.22 & 5.48$\pm$12.71 & 46.79$\pm$24.93 & 9.39$\pm$4.71 & 19.51$\pm$0.25 & 42.88$\pm$0.00 & 42.67$\pm$0.50 & 53.08$\pm$1.27 \\
		\cmidrule{2-15}
		& Sonar & 50(2) & 1.08$\pm$0.00 & 0.00$\pm$0.00 & 0.00$\pm$0.00 & 1.18$\pm$1.46 & 0.53$\pm$0.40 & 0.25$\pm$0.46 & 0.87$\pm$0.24 & 0.64$\pm$1.01 & \fontseries{b}\selectfont{1.98$\pm$0.00} & 0.88$\pm$0.00 & 0.78$\pm$0.46 & 0.00$\pm$0.00 \\
		& & 95(17) & 0.88$\pm$0.00 & 2.14$\pm$1.52 & 0.88$\pm$1.18 & 1.49$\pm$1.81 & 0.73$\pm$0.36 & 0.20$\pm$0.42 & 0.00$\pm$0.00 & 2.06$\pm$3.02 & 1.30$\pm$0.00 & 1.05$\pm$0.00 & 0.76$\pm$0.47 & \fontseries{b}\selectfont{2.15$\pm$2.02} \\
		& & 99(29) & 0.88$\pm$0.00 & 0.25$\pm$0.33 & \fontseries{b}\selectfont{1.55$\pm$1.54} & 0.98$\pm$1.39 & 0.73$\pm$0.42 & 0.23$\pm$0.39 & 0.81$\pm$0.34 & 1.55$\pm$2.93 & 0.61$\pm$0.00 & 1.24$\pm$0.00 & 1.05$\pm$1.01 & \fontseries{b}\selectfont{1.55$\pm$1.55} \\
		\cmidrule{2-15}
		& WDBC & 50(1) & 46.72$\pm$0.00 & 46.32$\pm$0.00 & 46.72$\pm$0.00 & 46.32$\pm$0.00 & 46.56$\pm$0.33 & 14.66$\pm$16.28 & 46.32$\pm$0.00 & 6.47$\pm$1.74 & - & 46.72$\pm$0.00 & 52.11$\pm$0.88 & \fontseries{b}\selectfont{52.21$\pm$0.00} \\
		& & 95(1) & 46.72$\pm$0.00 & 46.32$\pm$0.00 & 46.72$\pm$0.00 & 46.32$\pm$0.00 & 46.56$\pm$0.33 & 14.66$\pm$16.28 & 46.32$\pm$0.00 & 6.47$\pm$1.74 & - & 46.72$\pm$0.00 & 52.11$\pm$0.88 & \fontseries{b}\selectfont{52.21$\pm$0.00} \\
		& & 99(2) & 46.72$\pm$0.00 & 47.39$\pm$0.00 & 45.77$\pm$0.00 & 46.98$\pm$0.00 & 46.32$\pm$0.37 & 8.18$\pm$16.24 & 45.92$\pm$0.00 & 5.13$\pm$2.04 & 0.01$\pm$0.00 & 46.72$\pm$0.00 & \fontseries{b}\selectfont{52.66$\pm$0.67} & 49.19$\pm$0.00 \\       			  
		\cmidrule{2-15}
		& Aust & 50(1) & 3.01$\pm$0.00 & 3.01$\pm$0.00 & 3.01$\pm$0.00 & 3.98$\pm$1.75 & 3.01$\pm$0.00 & 0.30$\pm$0.95 & 2.55$\pm$0.00 & 1.85$\pm$0.54 & - & 3.44$\pm$0.00 & \fontseries{b}\selectfont{11.48$\pm$0.27} & 6.13$\pm$0.00 \\
		& & 95(1) & 3.01$\pm$0.00 & 3.01$\pm$0.00 & 3.01$\pm$0.00 & 3.98$\pm$1.75 & 3.01$\pm$0.00 & 0.30$\pm$0.95 & 2.55$\pm$0.00 & 1.85$\pm$0.54 & - & 3.44$\pm$0.00 & \fontseries{b}\selectfont{11.48$\pm$0.27} & 6.13$\pm$0.00 \\
		& & 99(1) & 3.01$\pm$0.00 & 3.01$\pm$0.00 & 3.01$\pm$0.00 & 3.98$\pm$1.75 & 3.01$\pm$0.00 & 0.30$\pm$0.95 & 2.55$\pm$0.00 & 1.85$\pm$0.54 & - & 3.44$\pm$0.00 & \fontseries{b}\selectfont{11.48$\pm$0.27} & 6.13$\pm$0.00 \\
		\cmidrule{2-15}
		& Segment & 50(2) & 38.42$\pm$0.00 & 31.72$\pm$0.70 & 36.73$\pm$0.91 & 31.34$\pm$0.20 & \fontseries{b}\selectfont{48.21$\pm$7.39} & 2.78$\pm$3.11 & 47.41$\pm$3.49 & 36.39$\pm$18.52 & 45.44$\pm$1.75 & 42.68$\pm$0.00 & 36.57$\pm$6.23 & 36.16$\pm$0.95 \\
		& & 95(4) & 47.25$\pm$2.75 & 54.72$\pm$2.61 & 51.38$\pm$0.00 & 54.50$\pm$2.68 & 45.11$\pm$7.32 & 2.46$\pm$7.00 & 54.62$\pm$3.14 & 42.62$\pm$7.24 & 45.07$\pm$0.29 & 48.87$\pm$0.00 & 49.13$\pm$9.28 & \fontseries{b}\selectfont{55.48$\pm$1.44}  \\
		& & 99(6) & 50.14$\pm$2.14 & 35.04$\pm$1.92 & 60.81$\pm$0.00 & 36.83$\pm$0.76 & 44.39$\pm$6.12 & 2.84$\pm$5.71 & 60.42$\pm$4.78 & 42.81$\pm$2.41 & 45.15$\pm$0.29 & 53.48$\pm$0.00 & 47.48$\pm$8.52 & \fontseries{b}\selectfont{61.64$\pm$1.66} \\
		\cmidrule{2-15}
		& BC & 50(1) & 73.47$\pm$0.00 & 72.38$\pm$0.00 & 73.47$\pm$0.00 & 71.12$\pm$2.08 & 73.16$\pm$0.69 & 0.00$\pm$0.00 & 75.46$\pm$0.00 & 63.42$\pm$15.24 & - & 75.58$\pm$0.00 & 77.23$\pm$0.59 & \fontseries{b}\selectfont{80.55$\pm$0.84} \\
		& & 95(7) & 73.00$\pm$0.00 & 69.80$\pm$0.00 & \fontseries{b}\selectfont{82.41$\pm$0.00} & 73.06$\pm$1.73 & 74.05$\pm$1.11 & 0.00$\pm$0.00 & 69.01$\pm$0.00 & 77.13$\pm$2.35 & 75.58$\pm$0.00 & 74.92$\pm$0.00 & 78.15$\pm$1.67 & 57.77$\pm$22.33  \\
		& & 99(9) & 73.00$\pm$0.00 & 52.16$\pm$12.07 & 73.47$\pm$0.00 & 47.05$\pm$25.77 & 73.51$\pm$0.73 & 0.00$\pm$0.00 & 70.97$\pm$0.00 & 72.33$\pm$9.60 & 74.78$\pm$0.00 & 74.92$\pm$0.00 & 77.32$\pm$1.31 & \fontseries{b}\selectfont{78.81$\pm$0.63} \\
		\cmidrule{2-15}
		& CMC & 50(1) & 1.49$\pm$0.00 & \fontseries{b}\selectfont{3.58$\pm$0.00} & 1.49$\pm$0.00 & 1.12$\pm$0.00 & 0.37$\pm$0.53 & 0.00$\pm$0.00 & 3.25$\pm$0.00 & 3.04$\pm$0.00 & - & 1.25$\pm$0.00 & 2.03$\pm$1.11 & 1.44$\pm$0.00 \\
		& & 95(2) & 1.39$\pm$0.00 & 0.71$\pm$0.00 & 0.00$\pm$0.00 & 0.67$\pm$0.00 & 1.10$\pm$0.67 & 0.00$\pm$0.00 & \fontseries{b}\selectfont{3.53$\pm$0.88} & 3.02$\pm$0.00 & 3.13$\pm$0.08 & 1.25$\pm$0.00 & 2.18$\pm$0.68 & 0.57$\pm$0.64  \\
		& & 99(5) & 1.38$\pm$0.00 & 0.00$\pm$0.00 & 0.00$\pm$0.00 & 0.00$\pm$0.00 & 1.31$\pm$0.72 & 0.00$\pm$0.00 & 1.58$\pm$0.00 & 2.91$\pm$0.45 & \fontseries{b}\selectfont{3.14$\pm$0.03} & 1.25$\pm$0.00 & 2.00$\pm$0.63 & 3.01$\pm$0.73 \\
		\cmidrule{2-15}
		& Yale & 50(3) & 58.48$\pm$0.84 & 58.61$\pm$1.52 & 61.11$\pm$0.87 & 65.48$\pm$2.02 & \fontseries{b}\selectfont{65.72$\pm$1.83} & 58.32$\pm$3.74 & 59.49$\pm$0.76 & 59.54$\pm$2.84 & 54.95$\pm$1.23 & 61.77$\pm$0.00 & 49.91$\pm$17.21 & 59.61$\pm$2.18 \\
		\bottomrule
	\end{tabular}
\end{adjustbox}
\end{table*}

\begin{figure*}[htb]
	\centering	
	\setlength\tabcolsep{1.5pt}	
	\begin{tabular}{ c  c c }			
		\includegraphics[scale=0.27]{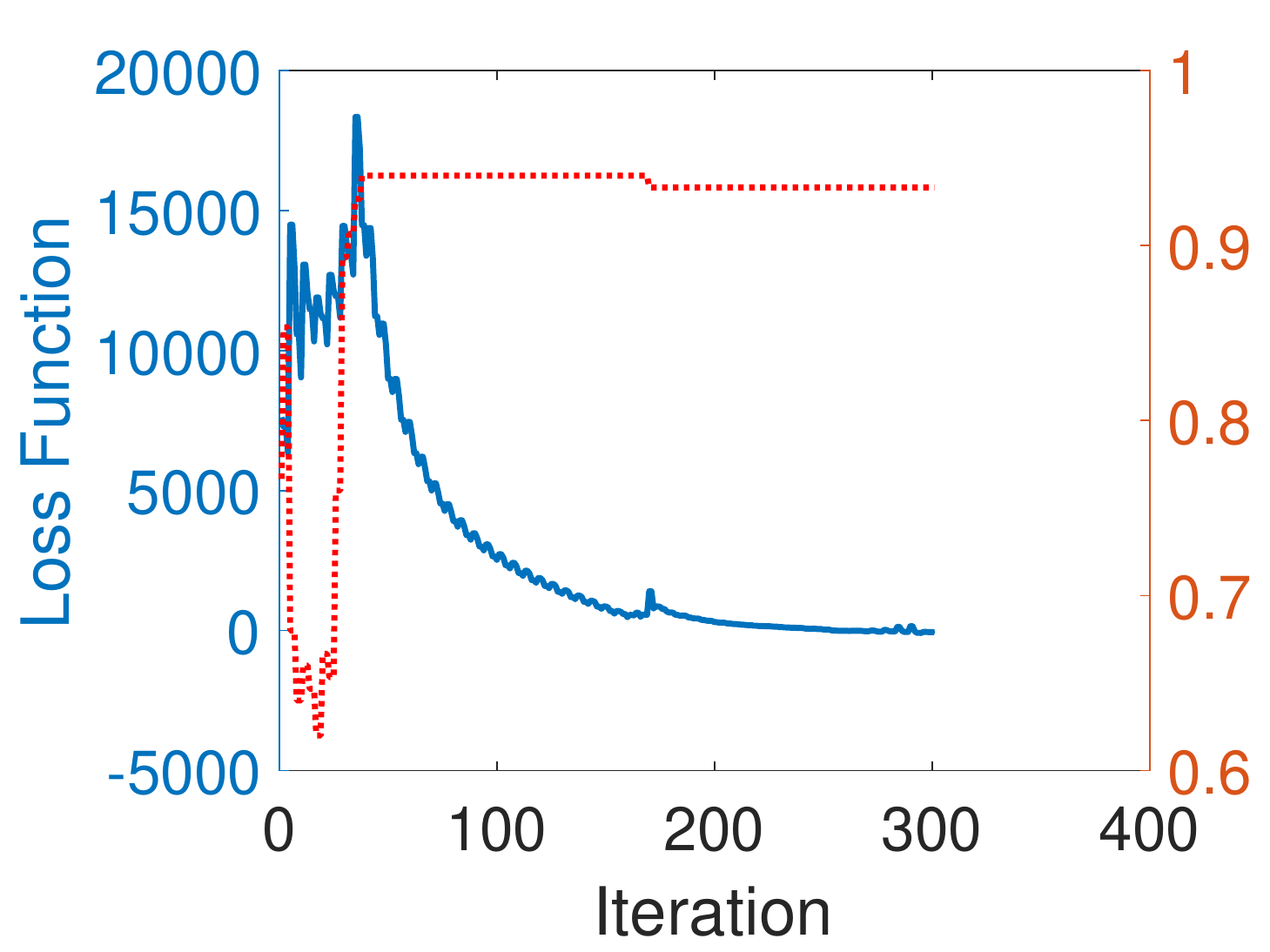} &\includegraphics[scale=0.27]{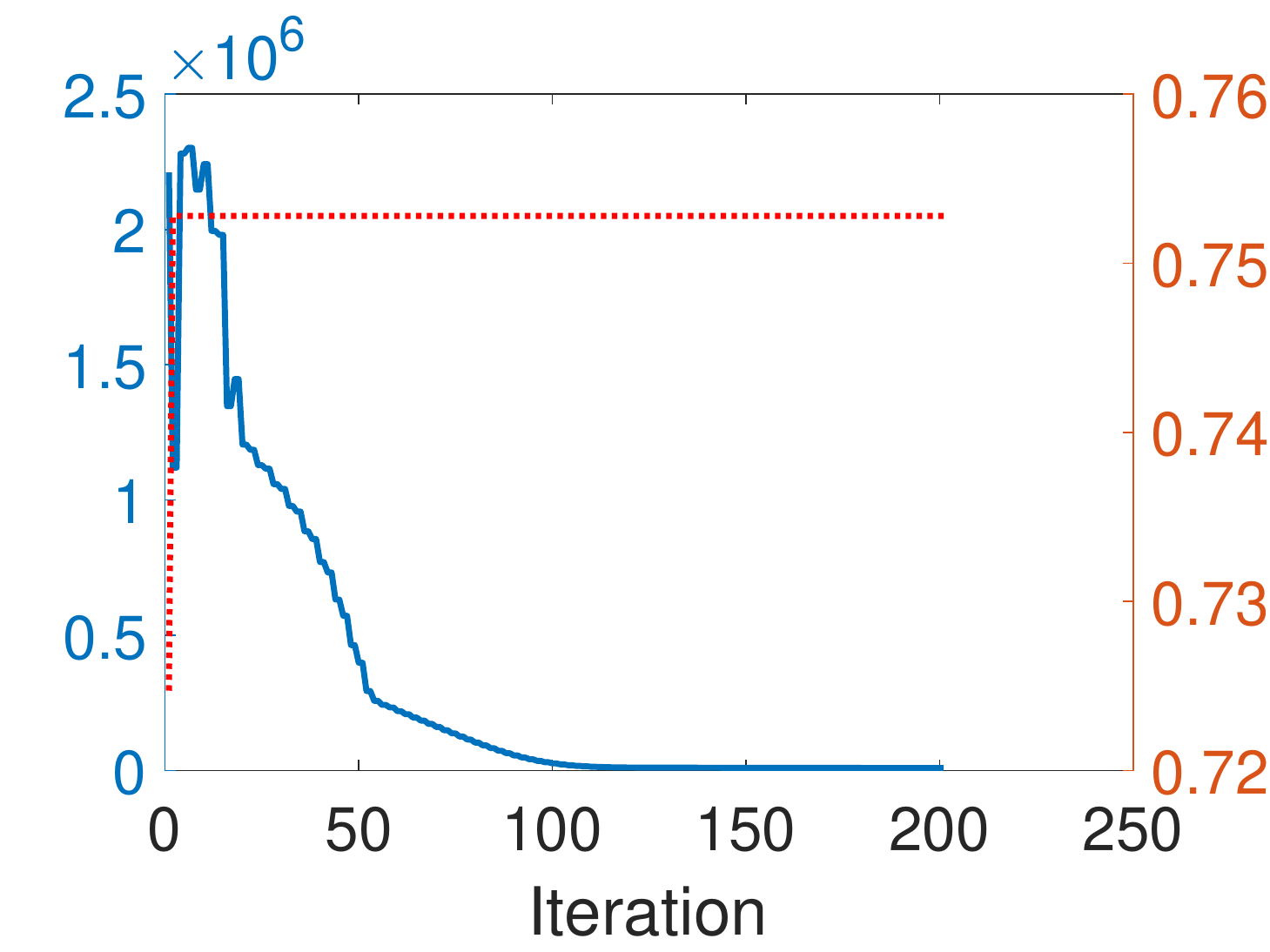} & \includegraphics[scale=0.27]{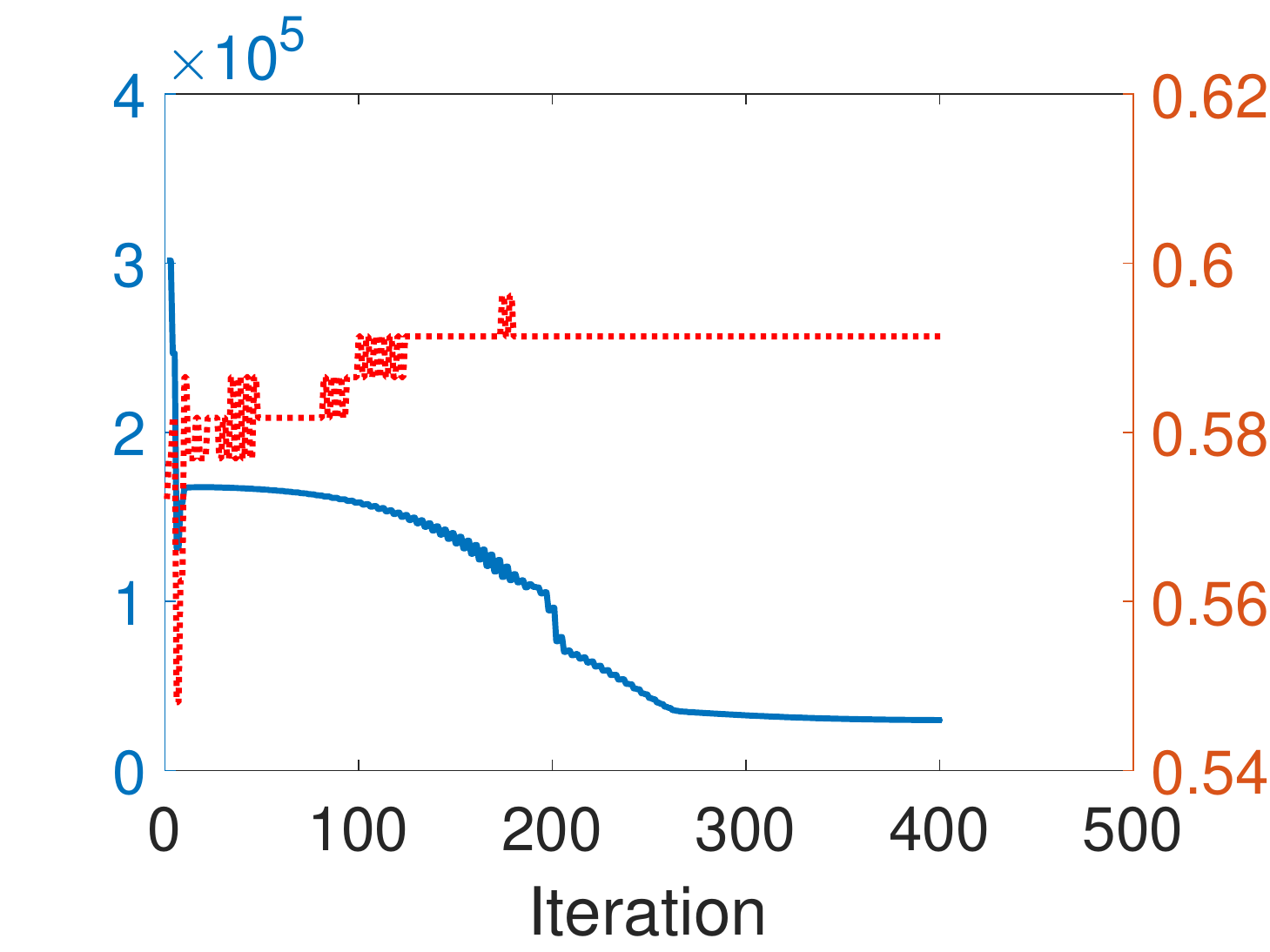}\\
		(a) Iris & (b) Wine & (c) Sonar \\
		\includegraphics[scale=0.27]{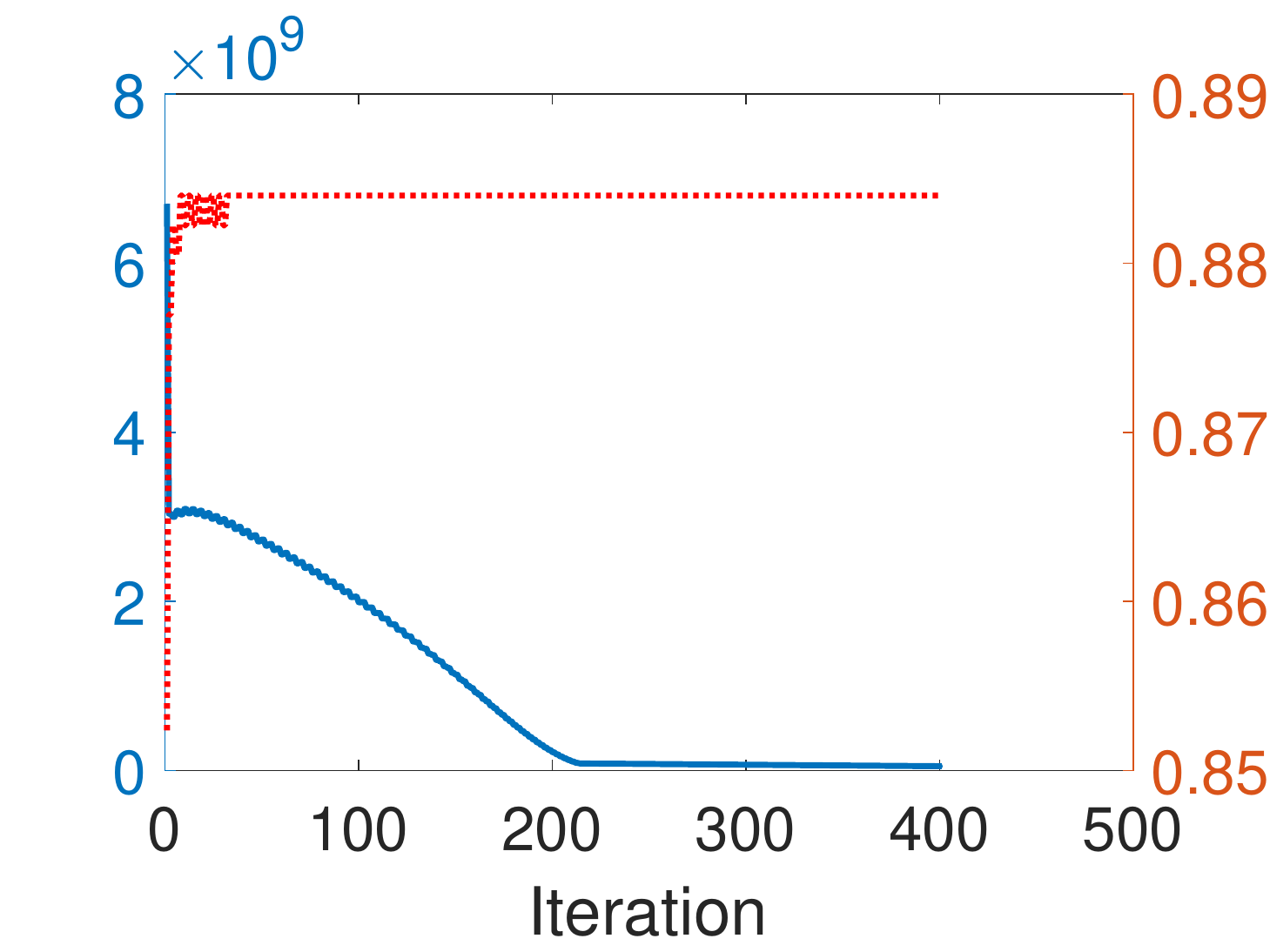}&\includegraphics[scale=0.27]{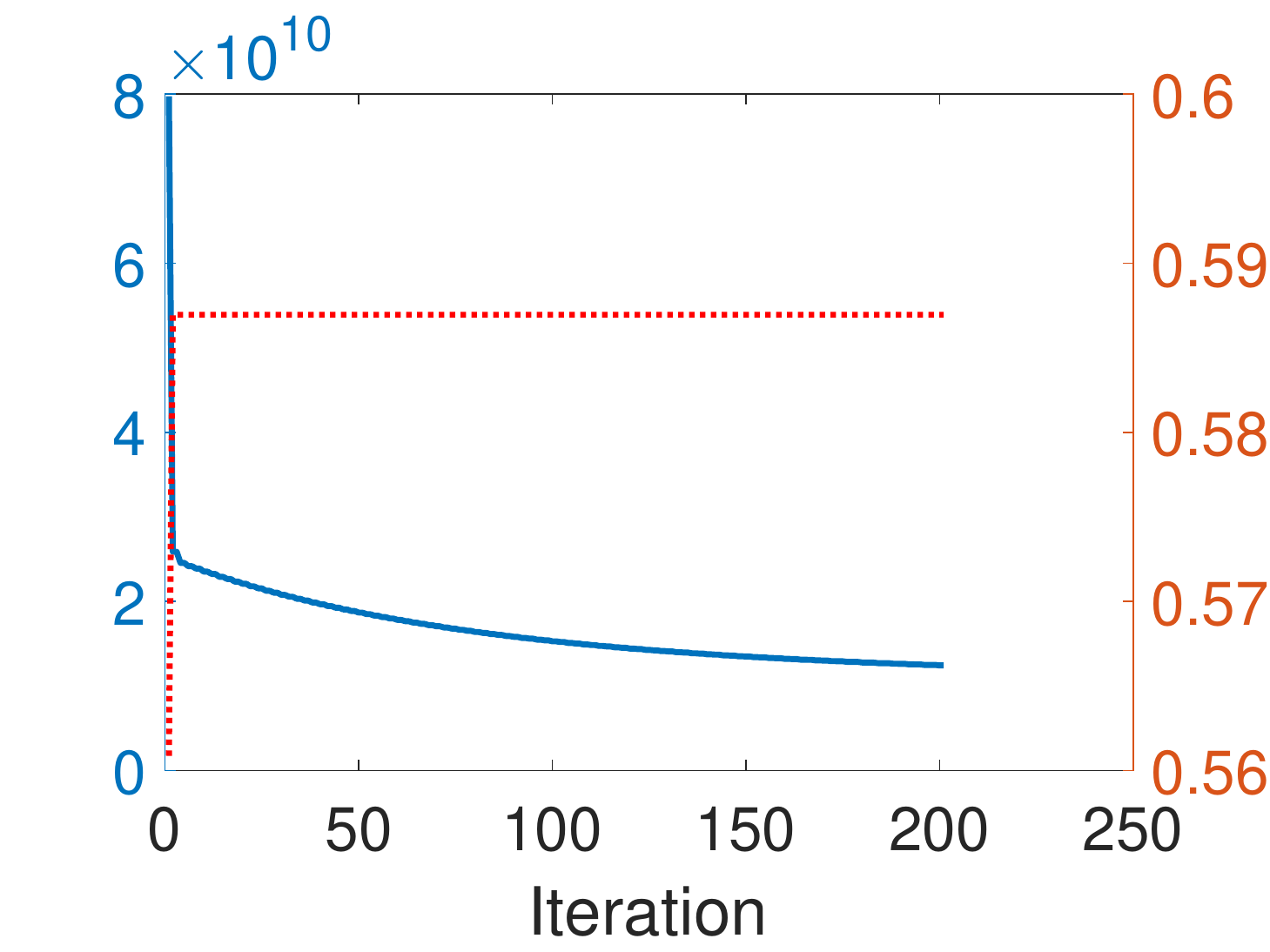} &\includegraphics[scale=0.27]{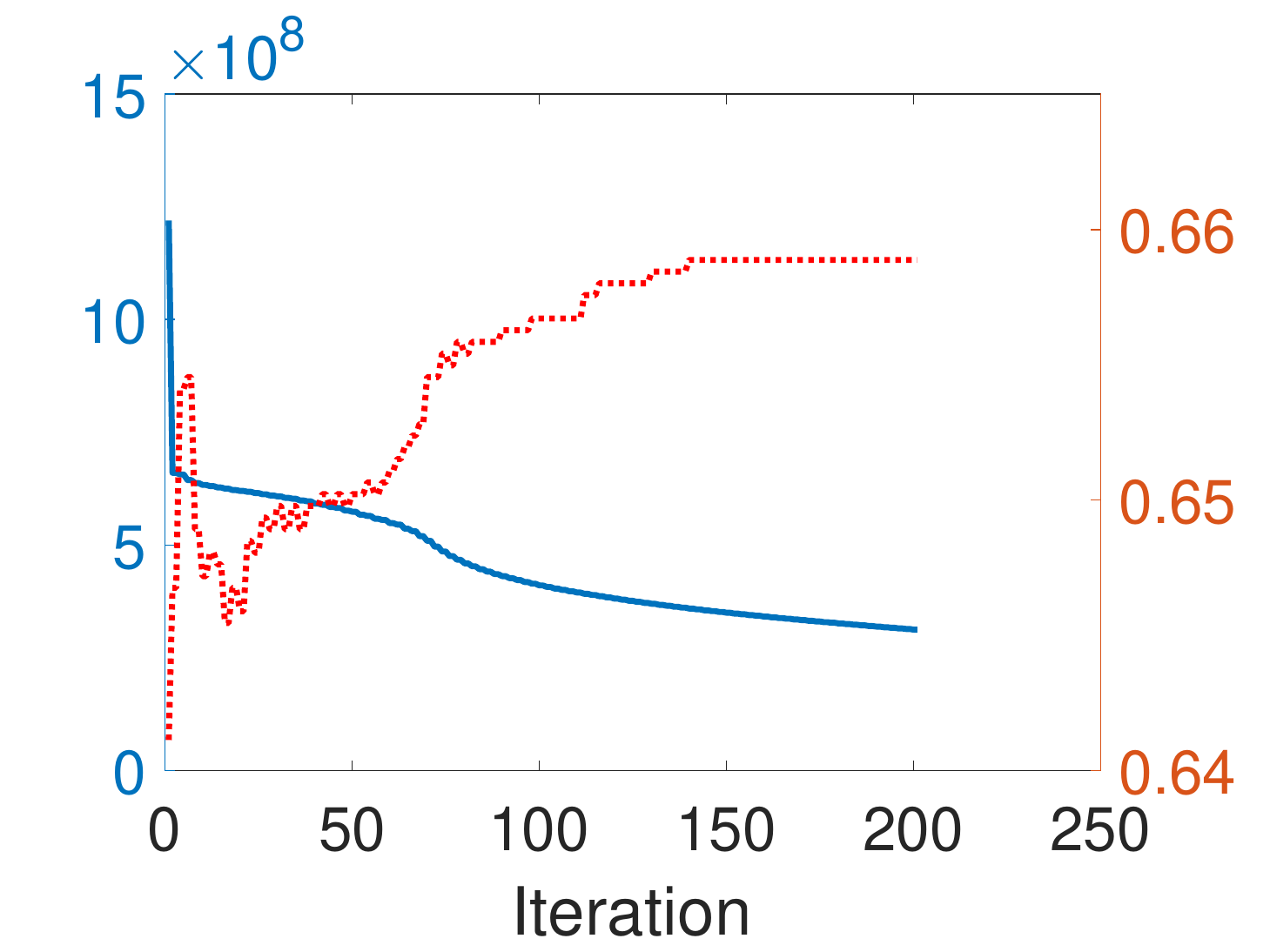}\\ 
		
		(d) WDBC & (e) Australian & (f) Segment\\ 
	\end{tabular}
	\caption{Clustering accuracy and the loss function in terms of iterations.}\footnotesize
	\label{figure:Convergence Behavior}
\end{figure*}

\subsection{Clustering Quality Comparison\label{sec:Accuracy Comparison}}

In this section, the accuracy and NMI results of different clustering methods are provided to demonstrate how SGP-MIC is able to improve the abilities of clustering. Here, each experiment is repeated 10 times and the averaged results with their standard deviation are considered. SGP-MIC algorithm is compared with PCA-FCM, ISOMAP-FCM, PPCA-FCM, FGPLVM-FCM, AE-FCM, and VAE-FCM. In these methods, first the low dimensional space is provided by PCA, ISOMAP, PPCA, FGP-LVM, Auto Encoder (AE) \cite{goodfellow2016deep}, and Variational AE (VAE) \cite{VAE} respectively in such a way that 50, 95, and 99 percent of data's variance are preserved. Then FCM is applied to cluster data in the new space. FGP-LVM is the fast version of GP-LVM using DTC approximation \cite{csato2002sparse,seeger2003fast}. In fuzzy c-means, each data point belongs to all clusters with different degrees of membership while in crisp clustering each data point only belongs to one cluster. The use of FCM allows achieving the higher performance in terms of clustering results, especially when the clusters are not well separated and they are overlapped \cite{budayan2009comparing}. Thus, we applied the FCM clustering as a baseline method after running different dimensionality reduction algorithms. SGP-MIC is also tested against AML, UNCA, DEC, SDC, and DEEPC methods where they do dimensionality reduction and clustering in a joint formulation. The results of AML and UNCA are the best averaged values obtained by tuning $\lambda$ among $[0.001, 0.01, 0.1, 1, 10, 100, 1000]$. The accuracy results of DEC are also computed by tuning $\lambda$ in $[6, 11, 16, 21, 26]$ and since $\lambda=6$ has the best average results on all data sets, the clustering values corresponding to this $\lambda$ are presented. These $\lambda$ values are suggested in their corresponding articles. The parameter tuning of SDC method is also the same as its article. We set network dimensions to $P$–64–32–$Q$ for all data sets, where $P$ is the data-space, and $Q$ is the latent-space dimensions. The other settings are the same as its original paper. The results are given in Table \ref{table:ACC comparison}. Second column defines the percentage of data's variances with their number of reduced dimension. In this table, the average and standard deviation of each experiment results are reported. Values in bold fonts are the highest averaged value on each row. By comparing the results, we can observe that SGP-MIC performs best in more set of experiments than other existing methods. Of the 22 different experiments, SGP-MIC has best accuracy and NMI in 7. The results show that there is not any relationship between the performance of the method and the number of samples or the number of features. Among all data sets, Segment and CMC data sets are the largest ones in term of number of data points and the results of our method are still good on them. Also, if we consider the number of features and we ignore data set Yale which is image data, the Sonar has more features than others, but the model is not ineffective on that. The main purpose of the model is to learn the low dimensional embedded space to be able to capture the main manifold structure of the data. Since we assumed different GPs on different features, the number of features/GPs does not affect the quality of the method and what is important is the correlation between samples for each feature. If some features could define good correlation between samples, then the model could learn a good latent embedded data. Here, the results on the image data set Yale is not so good. The reason is that in our model, we assume different features of each data set are independent while on image data sets there are lots of dependencies over the features on each local neighborhood. 

\subsection{Convergence Behavior\label{sec:Convergence Behavior}}

The convergence behavior of SGP-MIC method is assessed in this section. The values of KL-corrected bound and the clustering accuracy as functions of iterations on 6 UCI data sets are illustrated in Figure \ref{figure:Convergence Behavior}. This figure corresponds to a "one-shot" basis (i.e. each experiment was only run once with one setting of the random seed and the other parameters) while the low dimensional space is provided in such a way that 95 percent of data's variance is preserved. Each one of the iterations corresponds to the updating all parameters in the M-step or updating $q\left(\mathbf{S}\right)$ in the E-step. The loss function is the minus value of KL-corrected bound. By analyzing Figure \ref{figure:Convergence Behavior}, it is observed that in all cases, during the iterations, values of loss function decrease and the accuracy values increase. However, in very few cases, the values of loss function increase, which is due to update of $q\left(\mathbf{S}\right)$. As discussed before in Section \ref{sec:KLcorrected}, we use one approximation in updating $q\left(\mathbf{S}\right)$. Thus, after updating $q\left(\mathbf{S}\right)$ in the E-step, the loss function may increase, where rarely is observed along the curve.

\begin{figure}[t]
	\centering	
	\setlength\tabcolsep{1.5pt}		
	\begin{tabular}{ c  c }			
		\includegraphics[scale=0.31]{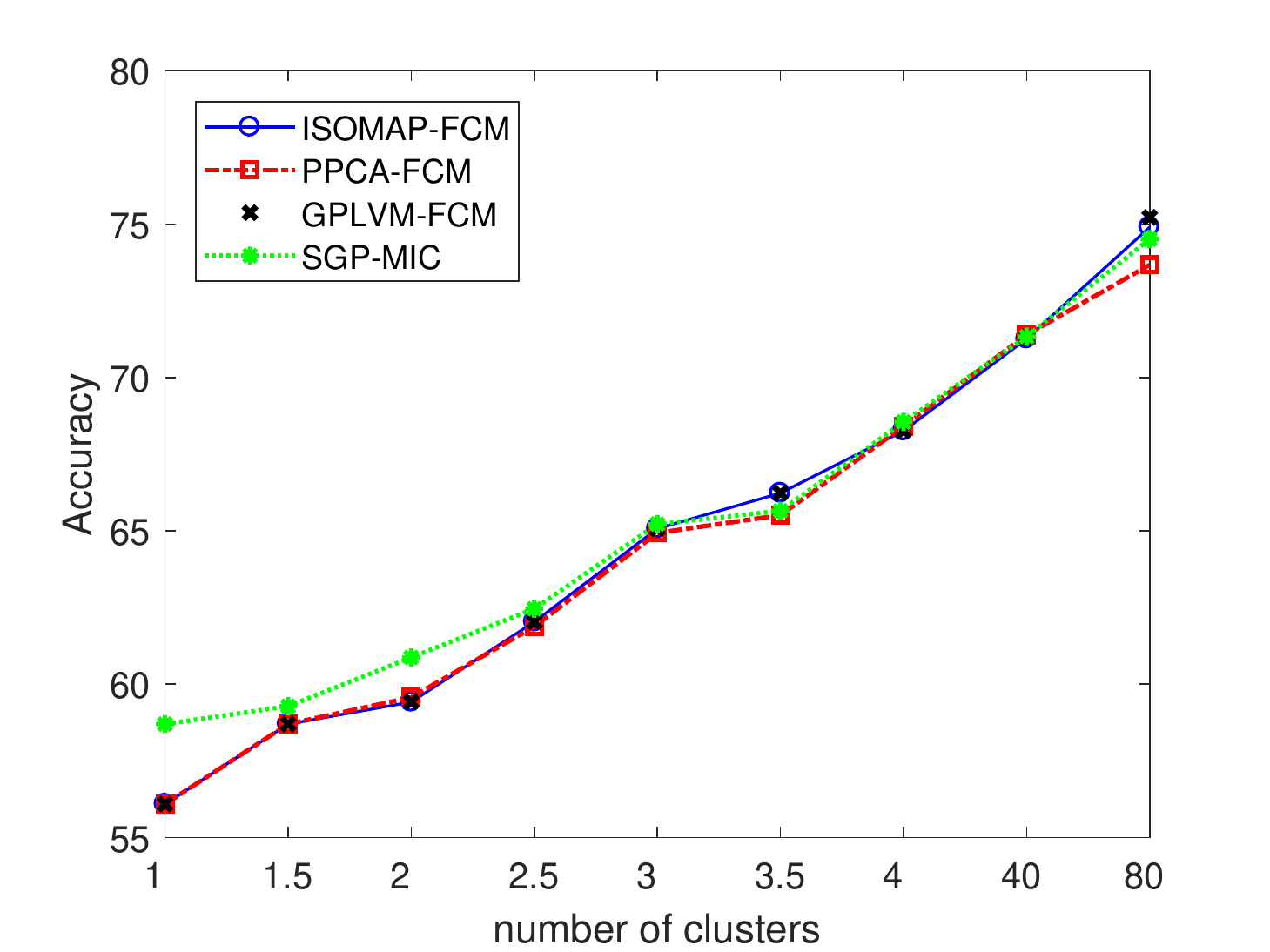}& \includegraphics[scale=0.31]{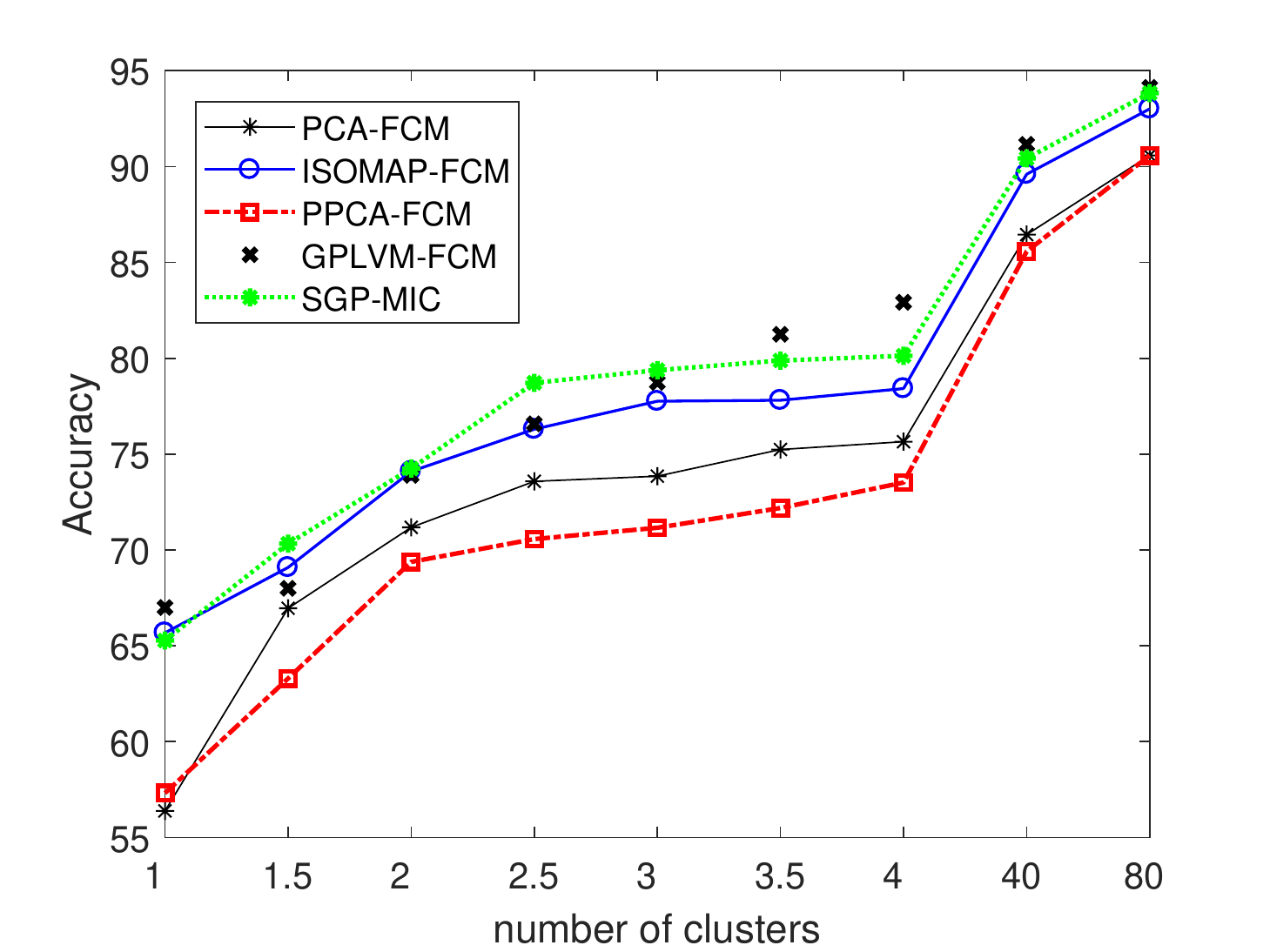} \\
		(a) Australian & (b) Segment\\ 
	\end{tabular}
	\caption{Variation of clustering accuracy with the number of clusters. (a) Australian, (b) Segment data sets.}\footnotesize
	\label{figure:Cluster Number}
\end{figure}

\begin{figure}[t]
	\centering
	\setlength\tabcolsep{1.5pt}	
	\begin{tabular}{ c  c c}			
		\includegraphics[scale=0.27]{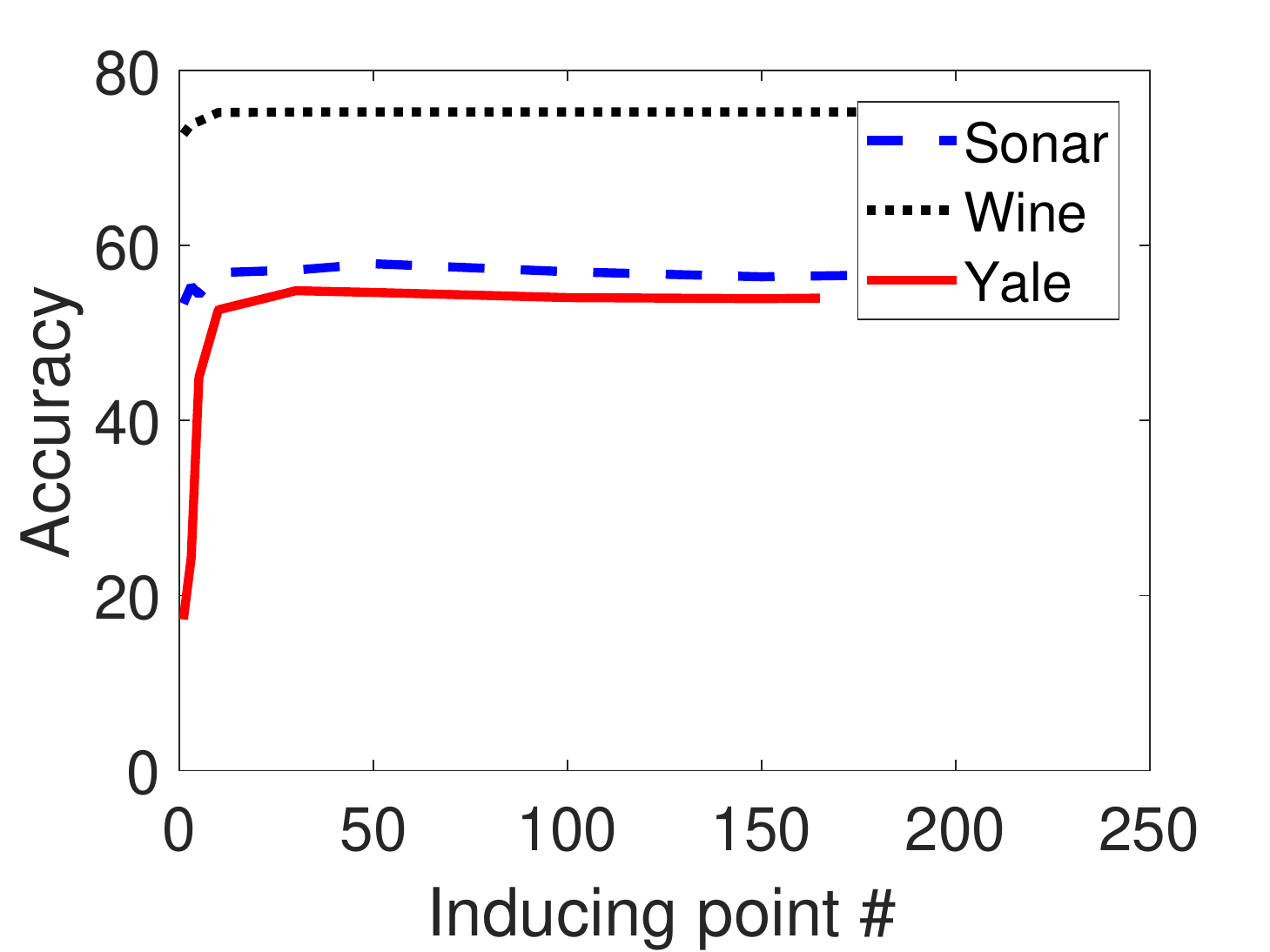} &\includegraphics[scale=0.27]{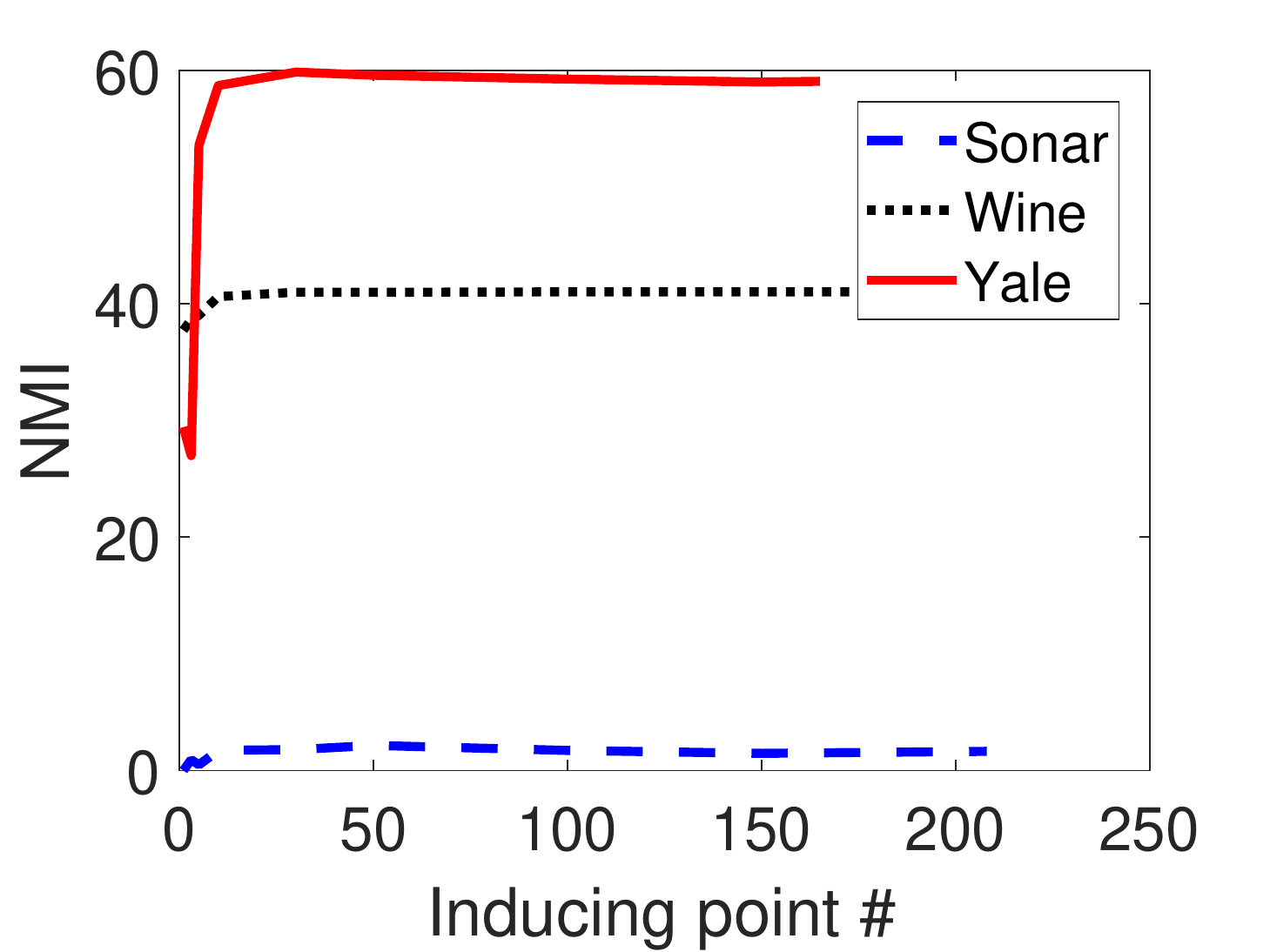} & \includegraphics[scale=0.27]{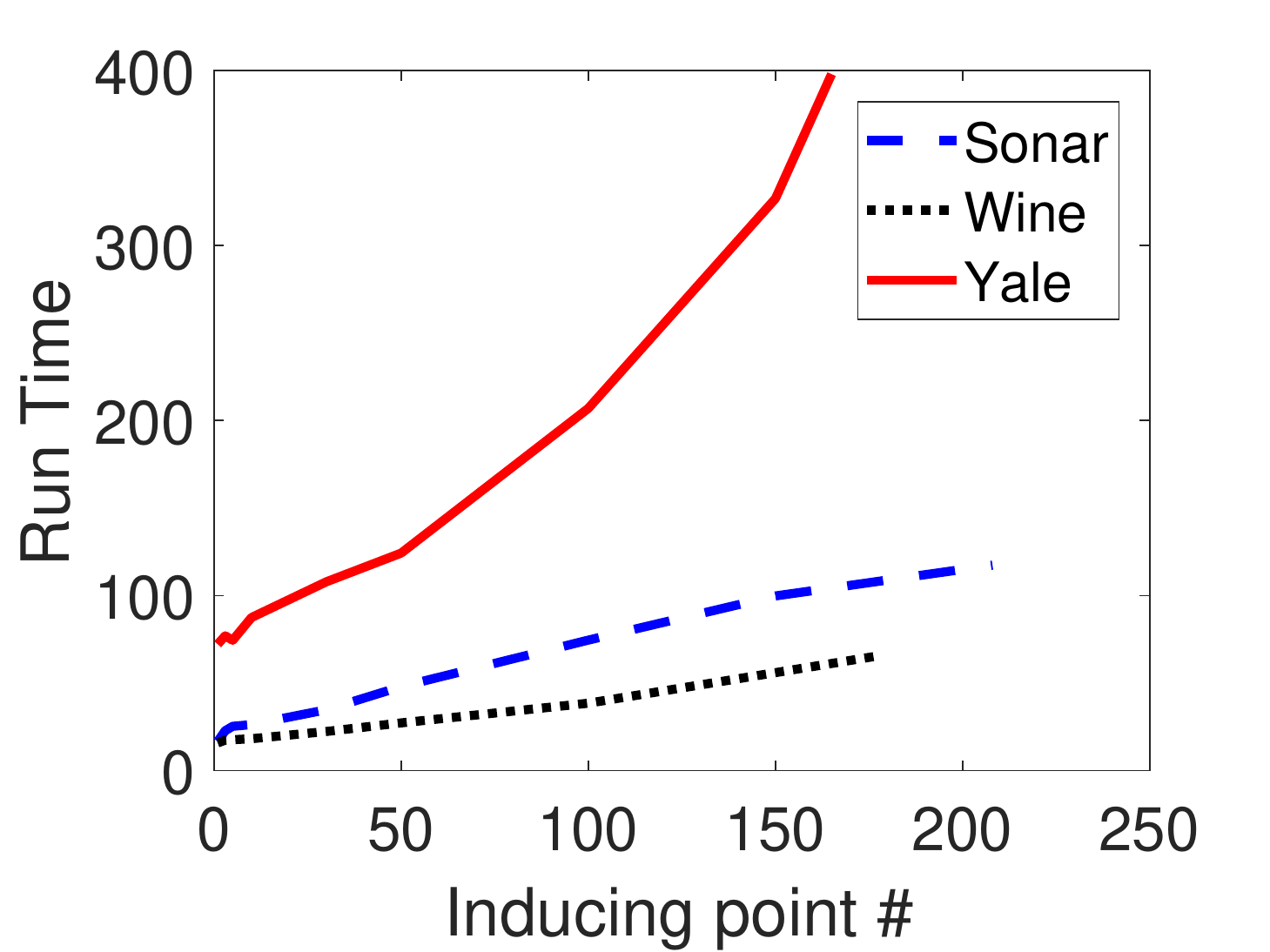}\\
	\end{tabular}
	\caption{Rates of accuracy, NMI and run time as $N'$ increases on Wine, Sonar and Yale data sets. The time measurement unit is second.\footnotesize}
	\label{figure:IndNum}
\end{figure} 

\subsection{Cluster Number Effect\label{sec:Cluster Number Effect}}

In SGP-MIC algorithm, we assume that we know the number of clusters. In this section, we assess the effect of cluster number in the results of SGP-MIC compared to other clustering algorithms. In this experiment, first the number of clusters is assumed to be equivalent to the number of classes and then it is increased to 80 times the number of classes. With a fixed cluster number, each of the SGP-MIC and other existing algorithms is repeated 10 trials and the averaged accuracies are computed. The accuracies as the functions of cluster number on two Australian and Segment data sets are shown in Figure \ref{figure:Cluster Number}. The low-dimensional space is provided by each algorithm in such a way that 95 percent of data's variance is preserved. By this figure, it can be seen that SGP-MIC outperforms other existing algorithms in different number of clusters. But, as the number of clusters increases, other algorithms might work better than SGP-MIC. By other algorithms, the low dimensional embedded space of input data is first obtained and then all data points have been clustered by the FCM clustering. Consider the case that the number of clusters is set to the number of data points. FCM could consider each data point as a unified cluster and gets 100\% accuracy result. Thus, it is clear that by increasing the number of clusters the performance of other algorithms would be increased. SGP-MIC learns both metric learning and clustering parameters together simultaneously. The clustering parameters are dependent to the metric learning parameters and vice versa. It is a tradeoff between two different objectives. When the cluster number is set to a rational value near to the number of classes, SGP-MIC has the best results compared to its counterparts without any joint formulation for clustering and metric learning, but when the cluster number is increased a lot, the growth rate of other methods could be more than the growth rate of our method.

\subsection{The Effect of Inducing points\label{sec:IndNumbEffect}}
In this section, we evaluate the performance and efficiency of SGP-MIC with respect to the number of inducing points $N'$, where its value is selected among $[1, 3, 5, 10, 30, 50, 100, 150, N]$. For each value, SGP-MIC is repeated 10 trials and the averaged accuracy, NMI, and run time are computed. The low-dimensional space is provided by 95 percent of data's variance. The results on Wine, Sonar, and Yale data sets are shown in Figure \ref{figure:IndNum}. As we see in this figure, when $N'$ increases, the accuracy, NMI and run time increase. What is interesting is that whenever $N'$ reaches to a fixed number, its increase does not effect on the clustering performance, although the run time is still increasing. This is why we set a fixed number for all of our experiments reported in Table \ref{table:ACC comparison}.

\section{Conclusion\label{sec:Conclusion}}

In this article, a new mixture model based on sparse Gaussian processes is proposed where integrates both dimensionality reduction and clustering in a joint formulation. Our approach is based on the dual interpretation of probabilistic PCA, which allows us to construct non-linear generalizations of the model and facilitates the introduction of additional constraints on the reduced dimensional representation. Moreover, using the sparse GP can help us to speed up the model. The traditional usage of Gaussian Processes is to model the stochastic process or time series, that is, the relation between temporal features. In this case, each sample is a time series data and the features of that sample are highly correlated, then we fit a GP on each sample. In our model, we consider that the features are independent but the samples are highly correlated. So, for each feature we fit a GP on all samples. Then we assume that the input space of that GP is latent and we learn this latent representation. If we would like to apply our method on some time series data, we need to fit two different GPs, one on the features and one of the samples. It could be a more complicated problem and we did not focus of that in this paper. Here, the main message of the paper is proposing a new method to learn the nonlinear latent representation of data using the sparse GP and at the same time learn the cluster of the data. Extension of this algorithm for handling time series data will be one of the future works. Learning the number of clusters and defining the number of desired dimensions in an automated sense will be also our next future research.

\section*{Acknowledgement\label{sec:Acknowledgement}}
We would like to express our deep gratitude to Professor Neil D. Lawrence and Professor Raquel Urtasun, for sharing their source codes published in GitHub\footnote{http://inverseprobability.com/mgplvm/}, where we apply them as a starting point of this research. Zahra also thanks Professor Neil D. Lawrence for hosting her in his research lab in 2016 at the Department of Neuroscience and Computer Science in University Sheffield. His kindness to give his time has been very much appreciated.

\bibliographystyle{elsarticle-num}
\bibliography{references}  

\section*{Appendix\label{sec:Appendix}}
\appendix
\section{Variational Posterior $q\left(s_{n,m}\right)$}
\label{appendix:qs}

By equating the functional derivative of Equation (\ref{eq:traditionalBound}) with respect to $q\left(s_{n,m}\right)$ to zero, we have
\begin{align}
	& \log q\left(s_{n,m}\right)=\nonumber \\
	& \sum_{i=1}^{P} \left\langle s_{n,m} \log \mathcal{N}\left(y_{n,i}|f_{n,i}^{\left(m\right)},\beta^{-1}\right)\right\rangle _{q\left(\mathbf{f}_{:,i}^{\left(m\right)}\right)}\nonumber \\
	& +s_{n,m}\log \mathcal{N}\left(\mathbf{x}_{n}|\bar{\mathbf{x}}_{m},\mathbf{C}_{m}\right) +s_{n,m}\log\pi_{m}+1,
\end{align}
where
\begin{align}
	& q\left(s_{n,m}\right)\propto\left( \right.\nonumber \\
	&\pi_{m}\mathcal{N}\left(\mathbf{x}_{n}|\bar{\mathbf{x}}_{m},\mathbf{C}_{m}\right) \nonumber\\
	& \left.\exp\left(\left\langle \log\prod_{i=1}^{P}\mathcal{N}\left(y_{n,i}|f_{n,i}^{\left(m\right)},\beta^{-1}\right)\right\rangle_{q\left(\mathbf{f}_{:,i}^{\left(m\right)}\right)}\right)\right)^{s_{n,m}}.
\end{align}
By multiplying a normalization constant such that $\sum_{m=1}^{M}q\left(s_{n,m}\right)=1$, the posterior distribution $q\left(s_{n,m}\right)$, proposed in Equation (\ref{eq:q_Snm}) is obtained.

\section{Variational Posterior $q\left(\mathbf{f}_{:,i}^{\left(m\right)}\right)$}
\label{appendix:qf}
First, we rewrite the first term of Equation (\ref{eq:traditionalBound}) in another form as follows:
\begin{align}
	& \sum_{n=1}^{N}\sum_{m=1}^{M}\sum_{i=1}^{P} \left\langle s_{n,m} \log \mathcal{N}\left(y_{n,i}|f_{n,i}^{\left(m\right)},\beta^{-1}\right)\right\rangle _{q\left(\mathbf{f}_{:,i}^{\left(m\right)}\right)q\left(s_{n,m}\right)}\nonumber \\
	& =\sum_{m=1}^{M}\sum_{i=1}^{P} \left\langle\log\prod_{n=1}^{N} \mathcal{N}\left(y_{n,i}|f_{n,i}^{\left(m\right)},\beta^{-1}\right)^{\left\langle s_{n,m}\right\rangle}\right\rangle _{q\left(\mathbf{f}_{:,i}^{\left(m\right)}\right)}\nonumber \\
	&
	=\sum_{m=1}^{M}\sum_{i=1}^{P} \left\langle \mathcal{N}\left(\mathbf{y}_{:,i}|\mathbf{f}_{:,i}^{\left(m\right)},\left(\mathbf{B}^{\left(m\right)}\right)^{-1}\right)\right\rangle _{q\left(\mathbf{f}_{:,i}^{\left(m\right)}\right)},
\end{align}
where $\mathbf{B}^{\left(m\right)}\in\Re^{N\times N}$ is a diagonal matrix with elements $b_{n,n}^{\left(m\right)}=\beta\left\langle s_{n,m}\right\rangle $.

Then, we substitute the above equation into the Equation (\ref{eq:traditionalBound}).
We take the functional derivative of this expression with respect to $q\left(\mathbf{f}_{:,i}^{\left(m\right)}\right)$ and equate
to zero, obtaining the following 
\begin{align}
	& \log q\left(\mathbf{f}_{:,i}^{\left(m\right)}\right)= \log\mathcal{N}\left(\mathbf{y}_{:,i}|\mathbf{f}_{:,i}^{\left(m\right)},\left(\mathbf{B}^{\left(m\right)}\right)^{-1}\right)\nonumber\\
	& +\log N\left(\mathbf{f}_{:,i}^{\left(m\right)}|\mathbf{0},\mathbf{K}_{fu}^{\left(m\right)}{\mathbf{K}_{uu}^{\left(m\right)}}^{-1}\mathbf{K}_{uf}^{\left(m\right)}\right)+1.
\end{align}

It is easy to see
\begin{equation}
	\begin{aligned}[b]
		q\left(\mathbf{f}_{:,i}^{\left(m\right)}\right) \propto \; &\mathcal{N}\left(\mathbf{f}_{:,i}^{\left(m\right)}|\mathbf{0},\mathbf{K}_{fu}^{\left(m\right)}{\mathbf{K}_{uu}^{\left(m\right)}}^{-1}\mathbf{K}_{uf}^{\left(m\right)}\right)\\
		& \mathcal{N}\left(\mathbf{y}_{:,i}|\mathbf{f}_{:,i}^{\left(m\right)},\left(\mathbf{B}^{\left(m\right)}\right)^{-1}\right).
	\end{aligned}
\end{equation}
\section{Positive Correction Term}
\label{appendix:CorrectionTerm}
The difference between two bounds (\ref{eq:traditionalBound1}) and (\ref{eq:KLBound1}) has a form of Kullback-Leibler divergence,
\begin{equation}
	\begin{aligned}[b]
		& \Delta\mathcal{L}\left(\mathbf{X},\boldsymbol{\theta},\beta\right) = \mathcal{L}_{\mbox{KL}}-\mathcal{L_{SV}} \\
		& =\sum_{i=1}^{P}\sum_{m=1}^{M}\int d\mathbf{f}_{:,i}^{\left(m\right)} q\left(\mathbf{f}_{:,i}^{\left(m\right)}\right) \\
		& \log\frac{q\left(\mathbf{f}_{:,i}^{\left(m\right)}\right)p\left(\mathbf{Y}|\left\langle \mathbf{S}\right\rangle ,\mathbf{X},\beta\right)}{\prod_{n=1}^{N}\mathcal{N}\left(y_{n,i}|f_{n,i}^{(m)},\beta\left\langle s_{n,m}\right\rangle \right)p\left(\mathbf{f}_{:,i}^{(m)}|\mathbf{X},\mathbf{X}_{u},\boldsymbol{\theta}\right)} \\
		& =\sum_{i=1}^{P}\sum_{m=1}^{M}\mbox{KL}\left(q\left(\mathbf{f}_{:,i}^{\left(m\right)}\right)|| \right. \\
		& \quad \quad\quad\quad \left.\frac{\prod_{n=1}^{N}\mathcal{N}\left(y_{n,i}|f_{n,i}^{(m)},\beta\left\langle s_{n,m}\right\rangle \right)p\left(\mathbf{f}_{:,i}^{(m)}|\mathbf{X},\mathbf{X}_{u},\boldsymbol{\theta}\right)}{p\left(\mathbf{Y}|\left\langle \mathbf{S}\right\rangle ,\mathbf{X},\beta\right)}\right) \\
		& =\sum_{i=1}^{P}\sum_{m=1}^{M}\mbox{KL}\left(q\left(\mathbf{f}_{:,i}^{\left(m\right)}\right)||p\left(\mathbf{f}_{:,i}^{\left(m\right)}|\mathbf{Y},\left\langle \mathbf{S}\right\rangle,\mathbf{X},\mathbf{X}_{u},\boldsymbol{\theta},\beta\right)\right).
	\end{aligned}
\end{equation}
This indicates that the new bound (\ref{eq:KLBound}) is still a lower bound on the log likelihood while it is an upper bound than standard variational bound.

\end{document}